\documentclass[twoside,11pt]{article}

\usepackage{blindtext}
\usepackage{cite}
\usepackage{amsmath,amssymb,amsfonts}
\usepackage{graphicx}
\usepackage{textcomp}
\usepackage{xcolor}
\usepackage{amsmath}
\usepackage{algorithm}
\usepackage{stfloats}
\usepackage{blindtext}
\usepackage{multicol,lipsum}
\usepackage{algpseudocode}
\usepackage{lipsum}
\usepackage{ulem}
\usepackage{amsmath}
\usepackage{amssymb}
\usepackage{mathtools}
\usepackage{amsthm}
\usepackage{comment}
\usepackage{lipsum}

\usepackage{jmlr2e}

\usepackage{lastpage}

\ShortHeadings{Data-Driv. Lear. Nonlin. Diff. Eqs using Func. Anal.}{S. Sh. Alaviani, Y. Qu, and G. W. Vogl}
\firstpageno{1}

\begin{document}

\title{Data-Driven Learning of Unknown Nonlinear Differential Equations Using Functional Analysis}


\author{\name Seyyed Shaho Alaviani  \email salavian@umn.edu \\
       \addr Department of Mechanical Engineering\\
       University of Minnesota-Twin Cities\\
       Minneapolis, MN 55455, USA
       \AND
       \name Yongzhi Qu \email yongzhi.qu@utah.edu \\
       \addr Department of Mechanical Engineering\\
       University of Utah\\
       Salt Lake City, UT 84112, USA
       \AND
       \name Gregory W. Vogl \email gregory.vogl@nist.gov \\
       \addr National Institute of Standards and Technology\\
       Gaithersburg, MD 20899, USA}

\editor{My editor}

\maketitle

\begin{abstract}
	In this paper, the problem of data-driven discovery of nonlinear ordinary differential equations (ODEs) is recast, and a new interpretable machine learning (ML) method is proposed. The proposed method aims to learn the unknown vector field of nonlinear dynamics without prior knowledge of the system's physics from only one single state trajectory's data. The proposed method has two fundamental differences with existing methods: 1) the formulation presented in this method is derived based on Functional Analysis and Operator Theory, and 2) the cost function is constructed in the function space as a distance between two functions as an integral, instead of the discrete-sum of errors used in existing ML approaches. An incremental learning algorithm is proposed to learn the unknown vector field to handle new training samples in an online manner. The proposed method can discover the unknown vector field from both forced and unforced autonomous and non-autonomous (or time-varying) dynamical systems. The proposed method is able to simultaneously discover unknown external forces as a function of time and unknown underlying dynamics. Finally, numerical examples are given to demonstrate the advantages of the proposed method. 
\end{abstract}


\section{Introduction}

Many real-world processes are governed by ordinary differential equations (ODEs) such as electrical circuits, plasma physics, epidemiology, neuroscience, medicine, and biological systems. Scientists have sought to discover mathematical models from data in the form of ODEs. This problem is referred to as \textit{system identification} (\cite{systemsidentificationSurvey2025} and \cite{nonlinearsystemident11111111}) or \textit{inverse problem} (\cite{inverseproblembook1}). The problem is also related to the \textit{network inference} (\cite{networkinferencedecember}) problem in biology that aims to model the interactions among different states of a system. As pioneering works, Johannes Kepler (1571-1630) discovered the laws of planetary motion from the planets’ trajectory data recorded by Tycho Brahe (1564-1601), and Isaac Newton (1642–1727) discovered the second law of motion. Once appropriate models are approximated, the models can serve various purposes such as control design, estimation, performance analysis, and prediction. Therefore, inverse problems are becoming very important in various scientific and engineering applications. 

In recent years, ODE discovery\footnote{ODE discovery is also referred to as \textit{global vector-field reconstruction}, a long-established problem in the field of nonlinear dynamics (see \cite{vectorfieldreconstruction}, \cite{vectorfieldreconstruction2} and references therein).} has become one of the major topics for research in science and engineering. Nonetheless, developing such models remains challenging and requires considerable effort from researchers since \textit{nonlinearity} is a key characteristic of dynamical systems observed in many scientific and engineering domains (see \cite{nonlinearsystemident11111111}). In this regard, a variety of notable methods and/or algorithms have been proposed to learn unknown dynamics such as sparse regression like Sparse Identification of Nonlinear Dynamics (SINDy) (\cite{sindy}), symbolic regression (\cite{symbolicregressionsurvey2024}),  neural ODEs (NODEs) (\cite{neuralODE}),  symbolic neural ODEs (\cite{interPNODE}, \cite{bayesianPNODE}), data-driven Koopman operator theory (\cite{koopman1}, \cite{koopman2}, \cite{wiliams}, \cite{koopmansurveyJanuary}, \cite{Koopmandecember1}), physics-informed neural networks (PINNs) (classic one (\cite{2}, \cite{PINNsurvey22}, \cite{PINNsurvey}) and improved ones (\cite{XTFC}, \cite{AIlorenz})), optimizing a discrete loss (ODIL) (\cite{ODIL}), structure-preserving machine learning methods (\cite{SPMLdecember1}, \cite{SPMLdecember2}), and operator learning such as DeepONet (\cite{deeponet}), Fourier neural operator (FNO) (\cite{FNO}), and Laplace neural operator (\cite{LaplaceNeuralOperator1}). A brief literature review of these methods is provided in Subsection 1.1 below.

Existing methods can be roughly divided into two categories: \textit{black box} or \textit{interpretable}\footnote{Despite interchangeable use of explainable ML and interpretable ML in the literature, there are some differences between them (see \cite{XAIandIAI} and \cite{interpretableMLLLLLLLL22222} for details).} models. In black-box models, the internal modeling process is opaque, providing no guidance to humans on which aspects of the information should be trusted or disregarded.
To address this challenge, researchers have developed the burgeoning field of \textit{interpretable machine learning (IML)} (\cite{interpretableMLLLLLLLL11111}, \cite{interpretableMLLLLLLLL22222}).

From another perspective, an intelligent system needs to incrementally update, accumulate, acquire, and utilize knowledge. This ability is known as \textit{incremental learning} or \textit{sequential learning} (\cite{tasksurvey6666666}, \cite{tasksurvey1111111}, \cite{tasksurvey2222222}). In recent years, this area of machine learning research has received much attention due to its potential capabilities in industrial and societal applications such as medical diagnosis, autonomous driving, and predicting financial markets (\cite{taskNature444444}). There are several problems for which incremental knowledge assimilation offers a solution: memory restrictions, data security/privacy restrictions, and sustainable artificial intelligence (\cite{tasksurvey333333333}). 

\begin{figure}[t!]
	\centering
	\includegraphics[scale=0.5]{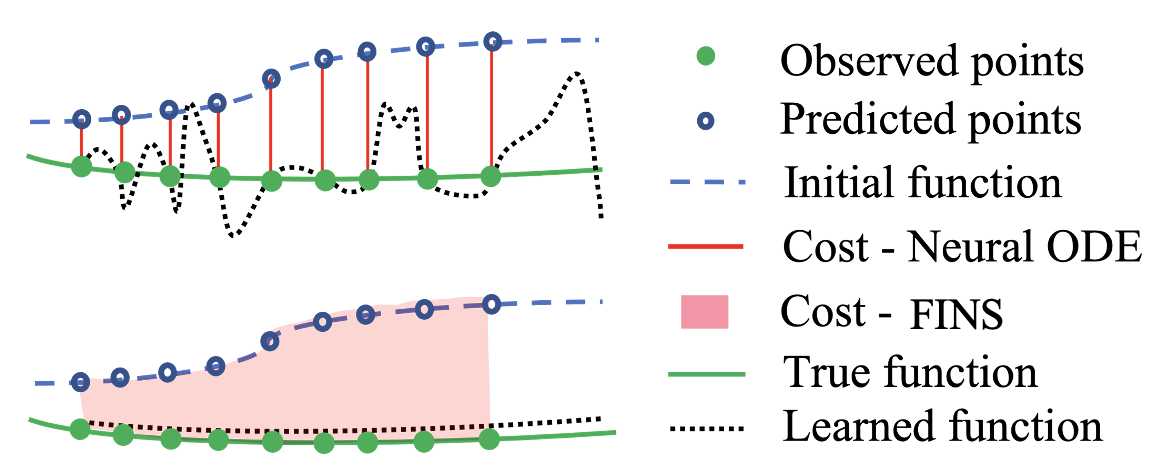}
	\caption{Difference between FINS and existing methods for data-driven dynamics learning, with NODEs as an example. (Top: Cost evaluation and potential learned function of NODEs; Bottom: Cost evaluation and potential learned function of FINS).}
	\label{fig1}
\end{figure}

In this paper, we recast the problem of nonlinear ODE discovery and present a new \textit{interpretable} ML method called FINS (\textbf{F}unctional-Analysis-Based \textbf{I}dentification of \textbf{N}onlinear \textbf{S}ystems). There are two fundamental differences between the proposed method and existing approaches: 1) we provide a learning framework which is based on Functional Analysis and Operator Theory (this allows developing an interpretable formulation\footnote{Unlike existing methods in the literature, we do \textit{not} use neural networks, lift to another space, solve an initial value problem with an ODE solver, do regression, or generate many ODEs and numerical solutions from random initial conditions.} with powerful generalization performance), and 2) the cost function is constructed in the function space as an integration of the distance between two functions instead of as a discrete-sum of errors, see Fig. \ref{fig1} for illustration. The merits of using the integration error are two folds. First, this allows to build machine learning models in the function space using the distance between functions to harvest relationships between training samples. From a machine learning perspective, the use of integration in the proposed method enables the modeling of temporal or sequential relationships between training samples, rather than treating them as unordered and uncorrelated samples using sum of (squared) errors. By integrating over time, the proposed method respects the continuity and smoothness inherent of systems' physics, thereby embedding the data in a more physically consistent learning framework. Second, in the case of learning ODEs, when a function represents the underlying vector field, the integration error allows direct estimation of integral error for an ODE solution without an explicit integral solver\footnote{Recently, a computationally efficient method based on a series of global integrals for each trajectory has been proposed in \cite{nodenosolver2025} for NODEs without relying on ODE solvers.}. This approach parallels the philosophy of Neural Ordinary Differential Equations (NODEs), which employ adaptive-step Runge–Kutta solvers to construct continuous-depth models. Our proposed method achieves a continuous representation by leveraging a Trapezoidal-rule-based numerical integration scheme. However, unlike NODEs, which often require an explicit integral solver, we directly evaluate the integral error in time and do not require an ODE solver in loop. In addition, an incremental learning algorithm, which can handle new training samples in an online manner is developed by using integration error. The proposed method can learn the nonlinear dynamics from only one single state trajectory’s data, without prior knowledge of the system’s physics, if the vector field is represented by finite degree polynomials\footnote{Application areas for polynomial vector fields can be exemplified as in models of HIV infection (\cite{HIV1}, \cite{HIV2}),  circadian rhythms in the chemistry of the eyes (\cite{circadianofEYE}), and population dynamics (\cite{populationdynamicsJanuary}).}. The results can be applied to both autonomous and non-autonomous (or time-varying) dynamical systems. Furthermore, this approach is applicable to learn the unknown underlying physics from both forced and unforced dynamical systems. An attribute of the proposed methodology is that it can simultaneously discover unknown external forces (as a function of time) and unknown underlying dynamics. The proposed framework can be used under state trajectory’s data and/or input-state trajectory’s data, that it is capable of learning the vector field from both zero and non-zero initial conditions (namely, it is \textit{initialization-free}).  The proposed framework offers a promising way to develop interpretable data-driven models for both complex natural phenomena and digital twins\footnote{The concept of digital twins involves the creation of virtual replicas of real-world processes.} of engineering systems. Despite the above advantages, a main limitation of this method is that it requires full state measurements or full input-state measurement and, hence, cannot give a representation of the vector field from input-output measurements.

The rest of the paper is organized as follows. In Section 2, preliminaries of Hammerstein operator and Hermite polynomials are given. In Section 3, the reformulation of ODE discovery and the incremental learning algorithm are provided. Finally, several numerical examples are presented in Section 4 in order to show the advantages of the proposed method, followed by conclusions and discussion in Section 5.

\subsection{Related Works}

\cite{Zadeh} prioritized the need for a system identification framework with an early overview performed by \cite{Astrom1971}. Since then, many researchers have paid much attention to data-driven learning of dynamical systems (\cite{ghadami}, \cite{Surveydecember1111}). In the following paragraphs, a brief review of existing methods is provided.

Dynamic Mode Decomposition (DMD) (\cite{dmd1}, \cite{dmd2}, \cite{dmd3}, \cite{robustDMD2}, \cite{schmid2022}) is a data-driven reduced-order modeling method that constructs the best fitted linear system for the \textit{flow map} of a nonlinear dynamical system. The Koopman operator transforms a finite dimensional nonlinear system into a lifted infinite dimensional linear system in the Hilbert space of observables (\cite{koopman1}, \cite{koopman2}, \cite{wiliams}, \cite{Koopmandecember1}, \cite{mezic2021}, \cite{koopmansurveyJanuary}). By selecting appropriate lifting functions known as \textit{embedding functions}, a linear representation of the nonlinear dynamics in the observable space can be approximated from data (\cite{wiliams}). However, pre-determining these embedding functions (also called a \textit{dictionary}) from data is still an open challenge although some researchers such as \cite{deepkoopman} and \cite{deepkoopman2222222} have tried to approximate embedding functions by deep learning methods. Similar to DMD, the Koopman operator acts on the \textit{flow map} of a nonlinear dynamical system. Note that relations between extended DMD (EDMD) and Koopman operator theory have been investigated in \cite{wiliams}. EDMD and Koopman operator with inputs have also been investigated (\cite{koopmanwithcontroldecemberKutz}, \cite{Koopmanwithcontrol22222222}, \cite{Koopmandecember1}\footnote{To apply the results in \cite{Koopmandecember1} (that are for autonomous systems) for the case where an input exists, a condition that the input be a constant between two snapshots (with uniform sampling time) needs to be satisfied. However, our results do not require the condition on the input.}). An extension of EDMD called  \textit{infinitesimal generator EDMD (gEDMD)} has been proposed in \cite{gEDMDDD}. For learning vector fields of deterministic systems, gEDMD reduces to SINDy introduced below.

A primary way to discover nonlinear vector fields is to fit coefficients of a linear combination of basis functions with training data (\cite{WWWWWWWangdecember1111}, \cite{WWWWWWWangdecember2222}). \cite{Wangcompressedsensing2011} showed how vector fields can be identified with the compressed sensing (\cite{Compressedsensing1}, \cite{Compressedsensing2}) approach using a dictionary of polynomials (further investigation with applications to networks is given in \cite{Wangcomp2021december2021}). By using the principles of sparse regression (\cite{sparseregression11}) and compressed sensing, as well as considering a larger dictionary of functions, SINDy (\cite{sindy}) has been an effective and foundational method for approximating unknown vector fields of dynamical systems in computational science and engineering. However, a common issue with SINDy is sensitivity to the approximation of numerical derivatives. Recently, robust versions of SINDy have been proposed (e.g., \cite{sindynoise}, \cite{Sindynoisedecember1111}, \cite{EnsembleSidny}, \cite{IntegralSindydecember011011}, \cite{Sindyrungekutta}) to overcome this limitation. SINDy has been also extended to SINDYc for systems with control input (\cite{sindyccc}). The success of SINDy relies on properly chosen candidate functions, which may not be enough to discover the underlying vector field. This opens a need for another regression problem introduced next.

Symbolic regression (SR) (\cite{symbolicregressionsurvey2024}), as a subfield of interpretable machine learning, requires little prior knowledge to search over mathematical expressions of a data set. Unlike conventional regression, the optimization task in SR is to minimize the loss function over a space of admissible models (or functions). Since the optimization needs to be solved over a discrete search space, derivatives do not exist. Moreover, the size of the search space in SR grows exponentially with model complexity as a NP-hard (non-deterministic polynomial-time hardness) problem (\cite{symbolic}). Like conventional regression, the success of SR relies on properly defined candidate functions (\cite{symbolic}). Besides the use of SR for modeling \textit{static} equations from data, SR has been applied for learning dynamical systems without input (\cite{symbolicdecember1}, \cite{symbolicdecember2}, \cite{symbolicdecember3}, \cite{symbolicdecember4}, \cite{symbolicdecember5}, \cite{symbolicdecember6}) where SR is applied for each individual scalar equation of the original dynamical system (that grow with increase of dimension of state). As such, SR for learning dynamical systems adds more difficulties and computational costs.

In cases where the basis functions or the observables (used in Koopman operators) are difficult to know a priori, the nonlinearities in the vector field may need to be estimated by universal approximators such as \textit{artificial neural networks (NNs)} (\cite{systemsidentificationSurvey2025}, \cite{nonlinearsystemident11111111}). Although NNs are usually \textit{black-box} techniques, their abilities in dealing with high-dimensional and nonlinear systems motivate scientists to use them for data-driven modeling of dynamical systems. Recurrent neural networks (RNNs)\footnote{RNNs use their internal state networks to store representations of recent inputs and reuse the outputs to process the time series.} can be used to model the vector field.  Different approaches for RNN-based dynamical systems modeling have been studied (\cite{RNNODE}, \cite{RNNNN222222}, \cite{recurrenttopology}). Although RNNs are very successful in other applications, their performance in dealing with time-series data measured from dynamical systems is not very promising, as shown by \cite{neuralODE}. Neural ODEs (NODEs)\footnote{Before the introduction of NODEs (\cite{neuralODE}), similar architectures had been proposed to learn dynamical systems (e.g., see \cite{similarstructure11111} and \cite{similarstructure22222}).} (\cite{neuralODE}) were introduced as a continuous-depth model instead of specifying a discrete sequence of hidden layers to discover dynamical systems from time series data. Improvements of NODE have been proposed as Augmented NODEs (\cite{augmentedneuralODE}) and second-order NODEs (SONODEs) (\cite{sonode}), where they aim to approximate the \textit{vector field as a NN}. Combining SINDy and NODE as \textit{neural SINDy} was proposed in \cite{SindyNODEdecember}\footnote{As a special case where Runge-Kutta 4 (RK4) method is used as an ODE solver, similar approach (called RK4-SINDy) has also been proposed (\cite{Sindyrungekutta}).} (a similar structure with monomial basis functions was proposed as Ordinary Differential Equation Network (ODENet) by \cite{ODENet}). Although NODEs demonstrate an attractive way to discover dynamical systems, NODEs show memory efficiencies, long training times, and suboptimal results; therefore, some improvements based on regularization techniques, variational formulation, or homotopy methods have been performed (\cite{improvedNODEdecember1},\cite{nodenosolver2025},\cite{improvedNODEdecember2}, \cite{improvedNODEdecember3}).

NODEs model the vector field as a NN, whereas \textit{neural flows} (\cite{Neuralflows}) model the flow map as a NN and avoid using ODE solvers utilized in NODEs. NODEs and neural flows model differential equations in the time domain by NNs, whereas Neural Laplace (\cite{NeuralLaplace}) models the solutions of differential equations in the Laplace domain. Recently, neural-network-based methods have been used for learning dynamical systems satisfying conservation laws (\cite{icml2023conservation}, \cite{conservationlaw2024111111111}), chaotic systems (\cite{icml2023teacher}), and scalar ODEs (\cite{icml2023transformer})\footnote{Note that \cite{icml2023transformer} can learn a scalar ODE (in symbolic form) only from one single trajectory of data while (randomly) generating a total of $>3M$ scalar, autonomous, non-linear, first-order ODEs.}. 

Neural operators are deep learning models between infinite-dimensional function spaces, unlike standard NNs which are maps between finite-dimensional spaces. Neural operators can also be used for inverse problems (\cite{scientificMLChapterdecember}). In this regard, DeepONet (which was built upon the work on operator learning of \cite{ChenandChen1995}) was the first operator learning framework proposed by \cite{deeponet}. Since then, several other neural operators have been proposed such as Laplace neural operator (\cite{LaplaceNeuralOperator1}), Fourier neural operator (FNO) (\cite{FNO}), wavelet neural operators (WNOs) (\cite{WaveletNeuralOperator}), transformers\footnote{Transformers (\cite{transformerssurvey}) have also been used for solving SR (see \cite{symbolicregressionsurvey2024} for details).} (as a special case of neural operators) (\cite{attention}, \cite{transformerssurvey}), graph neural operators (\cite{GraphNeuralOperatorAnandkumar}, \cite{Neuraloperator2023december}), and Koopman neural operator (\cite{koopmanneuraloperator0000000}), to name a few. Unlike previous data-driven methods that approximate unknown ODEs only from \textit{state} trajectory's data, neural operators approximate the flow map of the ODE from \textit{input-state}\footnote{By ``input-state'' data we mean that not only are samples of state trajectory needed but samples of the input are also required.} data over a bounded time interval by using the relevant universal approximation theorem for operators. Operator learning can also be utilized for ODE discovery (\cite{scientificMLChapterdecember}). Nevertheless, current neural operator approaches typically require on the order of a hundred\footnote{For example, $200$ trajectories of training data for several ODEs (see Supplementary Information of \cite{LaplaceNeuralOperator1}).} input-output pairs, to approximate solution operators associated with ODEs. \textit{We require to mention that since we use only one single trajectory of input-state data, it is considered as one training data for neural operators and is not enough for neural operators to be adequately trained.}

Methods based on NNs suffer from major problems as they are black-box, cannot be generalized
well outside of the training regions, demand extensive data, and contain a large number of free parameters. These issues are problematic when trying to apply standard neural ODEs to dynamical systems. To address some of these issues, symbolic neural ODEs\footnote{Symbolic neural ODEs are polynomial neural networks (\cite{polynomialNNdecember11111}) used in neural ODEs.} (\cite{interPNODE}, \cite{bayesianPNODE}) and neural SINDy (\cite{SindyNODEdecember}, and similar structure like ODENet (\cite{ODENet})) have been proposed. The symbolic neural ODEs and neural SINDy perform on autonomous nonlinear systems. Recently, as an interpretable method, fusing symbolic regression with NNs has been investigated to discover the vector field of autonomous nonlinear systems under sufficiently weak unscaled additive noise (\cite{symbolicRgeressionviaNN2023}). Note that although (\cite{interPNODE}, \cite{bayesianPNODE}, \cite{SindyNODEdecember}, \cite{symbolicRgeressionviaNN2023}, \cite{ODENet})) provide interpretable methods, the trainability of their NNs is highly dependent on network architectures and variable initialization as well as the step size of the gradient descent.

The above methods (except DeepONet) for autonomous (time-invariant) dynamical systems cannot be directly applied to non-autonomous (time-varying) systems. Therefore, the aforementioned methods have been extended to non-autonomous
systems via online DMD (\cite{onlinedmd}), multi-resolution DMD (\cite{multiresdmd}), and Koopman operator for non-autonomous dynamical systems with periodic and quasi-periodic time dependence (\cite{perioKoopman}). In \cite{timevarying1}, a non-autonomous Koopman operator is used where the dynamical system is assumed to be locally time-invariant at each local stencil. In \cite{timevarying2}\footnote{Similar method to \cite{timevarying2} was proposed in \cite{timevarying202411111111111} by using DMD as well.}, the authors transform learning of non-autonomous systems into learning of locally parameterized systems over a set of discrete-time instances. 
Deep Koopman learning has also been proposed in \cite{deepkoopman111111} that approximates nonlinear time-varying systems as linear time-varying discrete-time systems.

\section{Preliminaries}

\textit{Notations:} $\mathbb{R}$ denotes the set of all real numbers. $\mathbb{N}$ denotes the set of all natural numbers. $\mathbb{R}^{n}$ represents the set of $n$-dimensional real space. $\mathbb{R}^{n \times m}$ denotes the set of all real $n$ by $m$ matrices. $A$ as a subset of $B$ is represented by $A \subseteq B$. $\times$ denotes the Cartesian product. $\dot{x}(t):=\frac{d}{dt}x(t)$ denotes differentiation of the function $x(t)$ with respect to $t$. $\emptyset$ represents the empty set. $(.)^{T}$ denotes the transpose of a matrix or a vector. For vectors $x_{i}, i=1,\hdots,n,$ $col\{x_{1},x_{2},\hdots,x_{n}\}:=[x_{1}^{T},x_{2}^{T},\hdots,x_{n}^{T}]^{T}$. $\nabla f(x)$ represents gradient of a function $f(.)$ at $x$. $\textbf{0}_{n}$ denotes the vector of dimension $n \in \mathbb{N}$ whose elements are all zero. $\approx$ denotes approximation which means almost equal to. For any vector $z \in \mathbb{R}^{n}, \Vert z \Vert_{2}=\sqrt{z^{T}z}$, $\Vert z \Vert_{1}:=  \sum_{i=1}^{n} \vert z_{i} \vert $ where $\vert . \vert$ denotes absolute value, and $\Vert z \Vert_{0}:= \vert \{i | z_{i} \neq 0, i=1,\hdots, n\}  \vert_{card}$ where $\vert . \vert_{card}$ represents cardinality of a set.

\noindent
\textbf{Definition 1} (\cite{krasnoselski}): Let $\Omega \subseteq \mathbb{R}$. The operator $$\Psi(u(t)):=\int_{\Omega} K(t,y)f(y,u(y))dy$$
where $f:\mathbb{R} \times \mathbb{R}^{n} \longrightarrow \mathbb{R}^{n}$, $n \in \mathbb{N}$, is a \textit{known} nonlinear mapping and $K:\Omega \times \Omega \longrightarrow \mathbb{R}$ is a known kernel is called \textit{Hammerstein} operator. Another form of the Hammerstein operator can be given as $\Psi(u(t)):=\int_{t_{0}}^{t} K(t,y)f(y,u(y))dy.$

\noindent
\textbf{Definition 2:} The \textit{orthogonal} polynomials 
\begin{equation}\label{hermitepolyderivation}
	H_{i}(y):=(-1)^{i} e^{y^{2}}\frac{d^{i}}{dy^{i}} e^{-y^{2}}, \quad{} i \in \mathbb{N} \cup \{0\}, y \in \mathbb{R},
\end{equation}
in the Hilbert space $L^{2}_{\psi}(\mathbb{R},\mathbb{R})$ where $\psi(t):=e^{-t^{2}}$ with the inner product
\begin{equation}\label{weightinhermitedefinition}
	\langle y,z\rangle_{L^{2}_{\psi}}:=\int_{-\infty}^{+\infty} \psi(s)y(s)z(s) ds,
\end{equation}
and the norm $\Vert y \Vert_{L^{2}_{\psi}}:=\sqrt{\langle y,y \rangle}_{L^{2}_{\psi}}$ are called \textit{Hermite polynomials}. For example, $H_{0}(y) =1,H_{1}(y) =2y ,H_{2}(y) =4y^{2}-2,$ and $H_{3}(y) = 8y^{3}-12y.$

\section{Main Results}

\subsection{Problem Formulation} 

In this subsection, we recast the problem of discovering nonlinear differential equations and provide a formulation for data-driven learning of linear/nonlinear autonomous/non-autonomous dynamical systems. We show that with a partially-observed state trajectory, we are able to recover the hidden dynamics, which is the underlying vector field.

Since we know that the true system belongs to the class of ODEs, the model system will have the same structure. Consequently, we consider a dynamical system represented by the following ordinary differential equation:
\begin{equation}\label{1}
	\dot{x}(t)=f(t,x(t))+\mathcal{F}(t)
\end{equation}
where $x:=[x_{1},x_{2},\hdots,x_{n}]^{T} \in \mathbb{R}^{n}$ is the state, $f:\mathbb{R} \times \mathbb{R}^{n}\longrightarrow \mathbb{R}^{n}, n \in \mathbb{N}$, is the \textit{unknown} nonlinear function to be approximated, $\mathcal{F}:\mathbb{R} \longrightarrow \mathbb{R}^{n}$ is the external force or input (which may or may not exist, and can be sampled) where $\mathcal{F}:=[\mathcal{F}_{1},\mathcal{F}_{2},\hdots,\mathcal{F}_{n}]^{T}$, and $t$ represents time. The problem of data-driven learning of (\ref{1}) is to approximate the nonlinear function $f(.,.)$ under a given data set defined as 
\textbf{}\begin{align}
	&\mathcal{D}_{t}:=  \{(t_{i},x(t_{i}),\mathcal{F}(t_{i})) |0 \leq i \leq k(t),  t_{0} < t_{1} < \hdots < t_{k(t)},  t_{k(t)} \leqslant t,	 k(t) \in \mathbb{N} \cup \{0\}\} \label{dataset}
\end{align}
where $x(t_{i})$ is the sample of the state trajectory $x(t)$ at time $t_{i}$, $\mathcal{F}(t_{i})$ is the sample of the external force $\mathcal{F}(t)$ at time $t_{i}$, and it is possible that $\vert \mathcal{D}_{t} \vert \longrightarrow \infty$ as $t \longrightarrow \infty$ in which $\vert \mathcal{D}_{t} \vert$ represents cardinality of the data set $\mathcal{D}_{t}$ at time $t$. Note that the cardinality of $\mathcal{D}_{t}$, i.e., $\vert \mathcal{D}_{t} \vert,$ is non-decreasing with respect to $t$, namely $k(\hat{t}) \leq k(\tilde{t})$ whenever $\hat{t} \leq \tilde{t}$. It should be noted that the lengths of sampling intervals, namely $t_{i+1}-t_{i}$ in which $0 \leq i \leq k(t)-1,$ can be different in our formulation, i.e., the formulation includes \textit{uniform} and \textit{non-uniform} sampling time. For simplicity, we write $k(t)=k$ throughout this paper. \textit{We should mention that we use all data in $\mathcal{D}_{t}$ for training in our proposed method}. We call the interval $[t_{0},t]$ \textit{training time} over which the training data are sampled. We call the time interval $[t, \mathcal{T}]$ for some fixed $\mathcal{T} \in \mathbb{R}$ \textit{prediction time}.

It should be noted that second-order systems, say $\ddot{z}(t)=f(t,z(t),\dot{z}(t))$, can be represented in the form of (\ref{1}) by changing variable $y=\dot{z}$ and considering the new state vector as $x=[z^{T},y^{T}]^{T}$ if $\dot{z}(t)$ can be sampled (or estimated) and provided in the data set. One way to estimate $\dot{z}(t)$ is to use Euler discretization method.   

\noindent
\textbf{Assumption 1:} $\mathcal{D}_{t} \neq \emptyset, \forall \, t \geq t_{0}$.

Note that measurements in the data set require a sampling time while the theory is exact in the limit of a zero sampling time. Now, by using Assumption 1, we convert the ordinary differential equation (\ref{1}) into the following integral form 
\begin{equation}\label{2}
	x(t)=x(t_{0})+\int_{t_{0}}^{t} f(s,x(s)) ds+\int_{t_{0}}^{t} \mathcal{F}(s) ds.
\end{equation}

Let us equip the space of functions $f:\mathbb{R} \times \mathbb{R}^{n} \longrightarrow \mathbb{R}^{n}$ with the inner product
\begin{equation}\label{normf}
	\langle f_{1},f_{2} \rangle_{L^{2}_{\varphi}}:=\int_{\mathbb{R}^{n+1}} e^{-\underline{\textbf{x}}^{T} \underline{\textbf{x}}} f_{1}^{T}(\underline{\textbf{x}}) f_{2}(\underline{\textbf{x}}) d \underline{\textbf{x}}
\end{equation}  
and norm $\Vert f \Vert_{L^{2}_{\varphi}}:=\sqrt{\langle f,f \rangle}_{L^{2}_{\varphi}}$ such that $\Vert f \Vert_{L^{2}_{\varphi}} < \infty$ where $\underline{\textbf{x}}:=[x_{1},x_{2},\hdots,x_{n},t]^{T},$ $d \underline{\textbf{x}}:= d x_{1} d x_{2} \hdots d x_{n} d t$, and $\varphi(\underline{\textbf{x}}):=e^{-\underline{\textbf{x}}^{T}\underline{\textbf{x}}}$. This Hilbert space is called $L^{2}_{\varphi}(\mathbb{R} \times \mathbb{R}^{n},\mathbb{R}^{n})$.

Let us also equip the space of trajectories $x:[t_{0},\infty) \longrightarrow \mathbb{R}^{n}$ with the inner product
\begin{equation}\label{normx}
	\langle x_{1},x_{2} \rangle_{L^{2}_{w}}:=\int_{t_{0}}^{\infty} w(s)x_{1}^{T}(s)x_{2}(s) ds
\end{equation} 
and norm $\Vert x \Vert_{L^{2}_{w}}:=\sqrt{\langle x,x \rangle_{L^{2}_{w}}}$ such that $\Vert x \Vert_{L^{2}_{w}}< \infty$ where $w(s)>0$ for $s \in [t_{0},\infty)$ and $\int_{t_{0}}^{\infty} w(s) ds < \infty.$ This Hilbert space is called $L^{2}_{w}(\mathbb{R},\mathbb{R}^{n})$. Now we define the following operator
\begin{equation}\label{operatorT}
	T(f)(t):=\int_{t_{0}}^{t} f(s,x(s)) ds
\end{equation}
where  $x(s), t_{0} \leq s \leq t,$ is \textit{known}, $T:L^{2}_{\varphi}(\mathbb{R} \times \mathbb{R}^{n},\mathbb{R}^{n}) \longrightarrow L^{2}_{w}(\mathbb{R},\mathbb{R}^{n})$, and the operator $T$ \textit{acts} on $f(.,.)$.

\noindent
\textbf{Remark 1:} If data-driven learning of (\ref{1}) over a bounded-time interval is desired, namely $t \in [t_{0},\textbf{T}]$ where $\textbf{T} \in \mathbb{R}$ is fixed, then one can consider $w(t) \equiv 1.$ Otherwise, we need to pre-select the weight $w(t), t \in [t_{0},\infty),$ in (\ref{normx}) such that the partially observed $x(t), t \in [t_{0},\infty),$ belongs to $L^{2}_{w}(\mathbb{R},\mathbb{R}^{n})$. One may choose $w(t)=e^{-at}$ where $a \in \mathbb{R}, \, a>0$, if needed.

At the first glance, it seems that the operator $T(f)$ defined in (\ref{operatorT}) is similar to the Hammerstein operator (see Definition 1). However, the Hammerstein operator acts on $x(t)$ where $f(.,.)$ is known, whereas the operator $T(f)$ acts on $f(.,.)$ where $x(t)$ is known. The operator $T(f)$ addresses the inverse problem instead of the forward problem in classical nonlinear dynamics analysis. Moreover, the Hammerstein operator is in general a nonlinear operator while the operator $T(f)$ is \textit{always} linear, as shown in the following lemma.

\noindent
\textbf{Lemma 1:} The operator $T(f)(t)$ defined in (\ref{operatorT}) is linear.

\noindent
\textit{Proof}: See Appendix A.

To reconstruct the unknown mapping $f(.,.)$ exactly, we need to reconstruct $x:[t_{0},\infty) \longrightarrow \mathbb{R}^{n}$ exactly from the data set $\mathcal{D}_{t}$ defined in (\ref{dataset}) and then solve (\ref{2}). However, this is \textit{not} practical for two reasons: 1) the cardinality of the data set $\mathcal{D}_{t}$ may not be enough to reconstruct $x(t), t \in[t_{0},\infty)$ (e.g., based on few samples, we cannot reconstruct $x(t)$ over the entire interval $[t_{0},\infty)$ exactly), and 2) it is impossible to wait for a long time to store a large data set to reconstruct $x(t), t \in[t_{0},\infty)$ exactly. To address this issue, based on data set $\mathcal{D}_{t}$ at time $t$, we define the operator sequence $\{T_{k}\}_{k=0}^{\infty}$ defined below:
\begin{equation}\label{3}
	T_{k}(f)(\bar{t}):= \int_{t_{0}}^{\bar{t}} f(s,x(s)) ds, \quad{} t_{0} \leq \bar{t} \leq t_{k}, k \in \mathbb{N} \, \cup \, \{0\},
\end{equation}
where $T_{k}: L^{2}_{\varphi}(\mathbb{R} \times \mathbb{R}^{n}, \mathbb{R}^{n}) \longrightarrow L^{2}_{w}([t_{0},t_{k}], \mathbb{R}^{n})$, and the inner products of the spaces $L^{2}_{\varphi}$ and $L^{2}_{w}$ are defined in (\ref{normf}) and 
\begin{equation}\label{normxtruncated}
	\langle x_{1},x_{2} \rangle_{L^{2}_{w}}:=\int_{t_{0}}^{t_{k}} w(s)x_{1}^{T}(s)x_{2}(s) ds,
\end{equation}
respectively (see also Remark 1). Note that (\ref{normxtruncated}) is a truncated version of (\ref{normx}) since we have truncated the time interval $[t_{0},\infty)$ to $[t_{0},t_{k}]$ in (\ref{3}). Therefore, based on the data set $\mathcal{D}_{t}$, (\ref{2}), and (\ref{3}), data-driven nonlinear dynamical system learning from a single state trajectory's data (which is theoretically valid over infinite time) reduces to the following sequence of optimization problems:

\begin{figure*}[t!]\label{figureschematic}
	\centering \includegraphics[scale=.8]{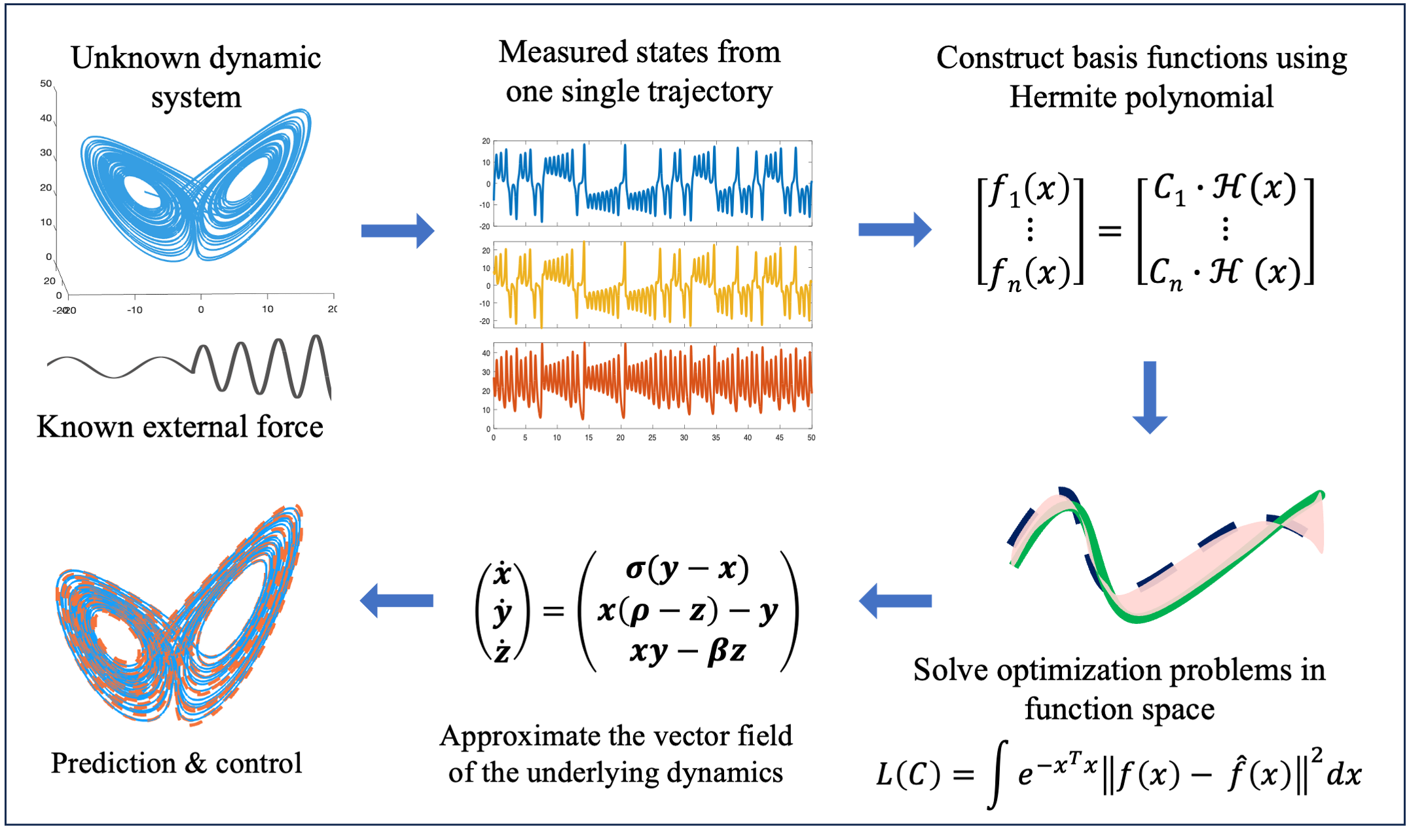}
	\caption{Schematic of FINS method.}
\end{figure*}

\begin{equation}\label{minimizationformulation}
	\begin{aligned}
		& \underset{f \in L^{2}_{\varphi}(\mathbb{R} \times \mathbb{R}^{n}, \mathbb{R}^{n})}{\text{min}}
		& & \Vert T_{k}(f)(\bar{t}) -x(\bar{t})+x(t_{0})+\int_{t_{0}}^{\bar{t}} \mathcal{F}(s)ds \Vert_{L^{2}_{w}} \\
		& \text{subject to}
		& & \{(t_{j},x(t_{j}),\mathcal{F}(t_{j}))\}_{j=0}^{k} \subseteq \mathcal{D}_{t}
	\end{aligned}
\end{equation}  
where $k \in \mathbb{N} \, \cup \, \{0\}$, and $L_{w}^{2}$ is defined in (\ref{normxtruncated}). In fact, based on Lemma 1, the optimization formulation (\ref{minimizationformulation}) is a least-squares problem over the function space $L^{2}_{\varphi}(\mathbb{R} \times \mathbb{R}^{n}, \mathbb{R}^{n})$ subject to partially known $x(s)$ and $\mathcal{F}(s)$, $s \in [t_{0},t],$ from the data set $\mathcal{D}_{t}$.

\noindent
\textbf{Remark 2:} Note that the formulation (\ref{minimizationformulation}) is also valid without external force, i.e., $\mathcal{F}(t) \equiv 0.$

\subsection{A Proposed Algorithm}

In this subsection, we provide an algorithm to approximate an optimal solution of (\ref{minimizationformulation}) by polynomials of finite degrees, based on the data set $\mathcal{D}_{t}$ defined in (\ref{dataset}). 

Any functions in $L^{2}_{\varphi}$ can be written as an infinite sum of Hermite polynomials (see Appendix B for details). Due to \textit{computational restrictions}, we require to approximate the function by a \textit{finite} sum of Hermite polynomials. Note that we choose Hermite polynomials as basis functions because they are orthogonal over the entire $\mathbb{R} \times \mathbb{R}^{n}$ and not over bounded domains. Let us \textit{pre-select} the highest degree of Hermite polynomials of the $i$\textsuperscript{th}, $i=1,\hdots,n+1,$ of $\underline{\textbf{x}}:=[x_{1},x_{2},\hdots,x_{n},t]^{T}$ as $N_{1}, N_{2}, \hdots,$ and $N_{n+1}$, respectively. We then have $N_{i}+1$ unknown coefficients for each $i$\textsuperscript{th} element of $\underline{\textbf{x}}$ (see Appendix B for details). Consequently, we define the matrix of coefficients $C \in  \mathbb{R}^{n \times (N_{1}+1)(N_{2}+1) \hdots (N_{n+1}+1)}$ and the finite degree Hermite polynomials $\mathcal{H}: \mathbb{R}^{n+1} \longrightarrow \mathbb{R}^{(N_{1}+1)(N_{2}+1) \hdots (N_{n+1}+1)}$ as follows:

\begin{equation}\label{matrixC}
	C:=
	\begin{pmatrix}
		c^{1}_{00\hdots0} & \hdots & c^{1}_{N_{1}N_{2} \hdots N_{n+1}} \\
		c^{2}_{00\hdots0} & \hdots & c^{2}_{N_{1}N_{2} \hdots N_{n+1}} \\
		\vdots & \ddots & \vdots \\
		c^{n}_{00\hdots0} & \hdots & c^{n}_{N_{1}N_{2} \hdots N_{n+1}} \\
	\end{pmatrix},
\end{equation}
\begin{equation}\label{matrixH}
	\mathcal{H}(\underline{\textbf{x}}):=\begin{pmatrix}
		H_{0}(x_{1})H_{0}(x_{2})\hdots H_{0}(x_{n})H_{0}(t)\\
  H_{0}(x_{1})H_{0}(x_{2})\hdots H_{0}(x_{n})H_{1}(t)\\
		\vdots \\
  H_{0}(x_{1})H_{0}(x_{2})\hdots H_{0}(x_{n})H_{N_{n+1}}(t)\\
  H_{0}(x_{1})H_{0}(x_{2})\hdots H_{1}(x_{n})H_{0}(t)\\
  H_{0}(x_{1})H_{0}(x_{2})\hdots H_{1}(x_{n})H_{1}(t)\\
  \vdots\\
		H_{N_{1}}(x_{1})H_{N_{2}}(x_{2})\hdots H_{N_{n}}(x_{n})H_{N_{n+1}}(t)	
	\end{pmatrix}.
\end{equation}

Now we provide the following proposition. 

\noindent
\textbf{Preposition 1:}
A solution of (\ref{minimizationformulation}) can be approximated by polynomials of finite degree at most $N_{1} N_{2} \hdots N_{n} N_{n+1}$, such that the higher order term is of the form $x_{1}^{N_{1}} \hdots x_{n}^{N_{n}} t^{N_{n+1}}$, from a solution of the following optimization
\begin{equation}\label{77777777777}
	\begin{aligned}
		& \underset{C_{k} }{\text{min}}
		& & \Vert \int_{t_{0}}^{\bar{t}} C_{k} \mathcal{H}(x(s),s) ds -x(\bar{t})+x(t_{0})+\int_{t_{0}}^{\bar{t}} \mathcal{F}(s)ds \Vert_{L^{2}_{w}}^{2} \\
		& \text{s.t.}
		& & \{(t_{j},x(t_{j}),\mathcal{F}(t_{j}))\}_{j=0}^{k} \subseteq \mathcal{D}_{t}
	\end{aligned}
\end{equation}
where $C_{k} \in \mathbb{R}^{n \times (N_{1}+1)(N_{2}+1) \hdots (N_{n+1}+1)}$,  $k \in \mathbb{N} \cup \{0\}$, and $N_{i}, i=1,\hdots,n+1$, is the highest degree of Hermite polynomials of the $i$\textsuperscript{th} element of $\underline{\textbf{x}}:=[x_{1},x_{2},\hdots,x_{n},t]^{T}$. 

\noindent
\textit{Proof}: See Appendix B.

Let us partition the matrix $C_{k}$ as $C_{k}=[C^{l}_{k}],l=1,2,\hdots,n,$ where $$C^{l^{T}}_{k}  \in  \mathbb{R}^{(N_{1}+1)(N_{2}+1) \hdots (N_{n+1}+1)}$$ and we define
\begin{equation}\label{mathcalKdefinition}
	\mathcal{K}(z):=\int_{t_{0}}^{z} \mathcal{H}(x(s),s) ds.
\end{equation}
Next, the main theorem in this paper is given.

\noindent
\textbf{Theorem 1:}
Solving (\ref{77777777777}) reduces to solving the following sequence of optimization problems:
\begin{equation}\label{888888888}
	\begin{aligned}
		& \underset{\mathcal{C}_{k}}{\text{min}}
		& & \mathcal{C}_{k}^{T} Q_{k} \mathcal{C}_{k} + P^{T}_{k} \mathcal{C}_{k}+U_{k} 
	\end{aligned}
\end{equation}
where $\mathcal{C}_{k}:=col\{C^{1^{T}}_{k},C^{2^{T}}_{k},\hdots,C^{n^{T}}_{k}\}$, 
\begin{align}
	&Q_{k}:= 	blockdiag \{\smash[b]{ \text{$n$ of}\int_{t_{0}}^{t_{k}} w(z) \mathcal{K}(z) \mathcal{K}^{T}(z) dz}\}, 
	\label{Qkdefinition}
\end{align}
\begin{align}
	P_{k}&:=col\{P_{k}^{1}, P_{k}^{2},\hdots, P_{k}^{n}\}, \label{Pkdefinition}\\
	U_{k}&:=\int_{t_{0}}^{t_{k}} w(z) \Vert x(t_{0})-x(z)+\int_{t_{0}}^{z} \mathcal{F}(s)ds \Vert_{2}^{2} dz, \label{Ukdefinition}
\end{align}
where
\begin{equation}\label{Pkldefinition}
	P_{k}^{l}:=2 \int_{t_{0}}^{t_{k}} w(z) \mathcal{K}(z) (x_{l}(t_{0})-x_{l}(z)+\int_{t_{0}}^{z} \mathcal{F}_{l}(s)ds)dz,
\end{equation}
and $Q_{k},P_{k},U_{k}$ are approximated based on the data set $\mathcal{D}_{t}$, and $ \mathcal{K}(z)$ is defined in (\ref{mathcalKdefinition}). 

\noindent
\textit{Proof}: See Appendix C.

\noindent
\textbf{Corollary 1:}
\textbf{(Convex case)} If $Q_{k}$, for some $k,$ is a positive semi-definite matrix, then an optimal solution of (\ref{888888888}) can be reached by solving $n$ linear algebraic equations, i.e.,
\begin{equation}\label{corollarylinearalgebraicequation}
	2\tilde{Q}_{k} C^{l^{*}}_{k}=-P^{l}_{k}, \quad{} l=1,2,\hdots,n,
\end{equation}
where 
\begin{align}
	\tilde{Q}_{k}&:=\int_{t_{0}}^{t_{k}} w(z) \mathcal{K}(z) \mathcal{K}^{T}(z) dz, \label{Qktildedefinition}	
\end{align}
$P_{k}^{l}$ is defined in (\ref{Pkldefinition}),  $\mathcal{C}^{*}_{k}:=col\{C^{1^{*^{T}}}_{k},C^{2^{*^{T}}}_{k},\hdots,C^{n^{*^{T}}}_{k}\}$ is an optimal solution of (\ref{888888888}), and the matrices $\tilde{Q}_{k}$ and $P^{l}_{k}, l=1,\hdots,n$ are approximated based on the data set $\mathcal{D}_{t}$, and $\mathcal{K}(z)$ is defined in (\ref{mathcalKdefinition}). 

\noindent
\textit{Proof}: See Appendix D.

\noindent
\textbf{Remark 3:}
\textbf{(Non-convex case)} If $Q_{k}$, for some $k,$ is \textit{not} a positive semi-definite matrix, then the algorithm in \cite{nonconvex} can be used to find an optimal solution of (\ref{888888888}). In this paper, such an algorithm for non-convex problems is not provided. Note that a solution of (\ref{corollarylinearalgebraicequation}) is a stationary (or critical) point of the cost function in (\ref{888888888}), which may be a sub-optimal solution for non-convex cases.

\textit{A challenge in Corollary 1 is to calculate the matrices $\tilde{Q}_{k}$ and $P^{l}_{k}, l=1,\hdots,n,$ over a long-time horizon by using data set $\mathcal{D}_{t}$.  In the rest of this subsection, we use a specific integral approximation to derive an incremental learning algorithm to address this challenge.}

\begin{algorithm}[b!]
	\caption{FINS}
	\label{algorithm1}
	
	\noindent \textbf{Input:}  Data set $\mathcal{D}_{t} \neq \emptyset$ , $N_{1}, N_{2}, \hdots, N_{n+1},$ $\mathcal{H}(\underline{\textbf{x}})$, and $w(t)$

	\textbf{Initialization:} $\tilde{Q}_{0}=\textbf{0}, \mathcal{K}_{0}=\textbf{0}, P_{0}^{l}=\textbf{0}, \theta^{l}_{0}=\textbf{0},l=1,\hdots,n.$
	
	\textbf{Output:} $\hat{f}(t,x)=$ approximation of unknown function $f(t,x)$ 
	
	\quad{}
	
	\textbf{IF} A datum is added to $\mathcal{D}_{t}$ at each $t_{k}$ \textbf{THEN} (\textit{incremental learning})	
	
	\textbf{FOR} $k=1$ \textbf{to} $\infty$ \textbf{DO}
	
	\textbf{\textbf{Step 1:}}	 
	
	Update $\tilde{Q}_{k}$ and $P_{k}^{l}$    \text{(defined in Appendix F)}

	\textbf{\textbf{Step 2:}} 
	
	Apply either Corollary 1 or Remark 3 (see also Subsection 3.2.1). 
	
	$\hat{f}_{k}(t,x) = C_{k}^{*} \mathcal{H}(x,t)$
	
	\textbf{END FOR}
	
	\textbf{END IF}
	
	\quad{}
	
	\textbf{IF} The cardinality of $\mathcal{D}_{t}$ is fixed for the training time \textbf{THEN}
	
	\textbf{DO} Step 1 for $k=1$ to $\vert \mathcal{D}_{t} \vert -1$.
	
	\textbf{DO} Step 2.		
	
	\textbf{END IF}

\end{algorithm}

\noindent
\textbf{Remark 4:}
Based on different methods to approximate the matrices $\tilde{Q}_{k}$ and $P^{l}_{k}, l=1,\hdots,n,$ by using the data set $\mathcal{D}_{t}$, different algorithms can be derived to solve (\ref{77777777777}). 

Now we derive an incremental learning algorithm to solve (\ref{corollarylinearalgebraicequation}) based on the data set $\mathcal{D}_{t}$ by using the following approximation of the integral of function $g(.)$:
\begin{equation}\label{integralapproximation}
	\int_{t_{k}}^{z} g(s) ds \approx \frac{g(t_{k})+g(t_{k+1})}{2} (z-t_{k}), \quad{} t_{k} \leq z \leq t_{k+1}.
\end{equation}
Note that the definite integration from $t_{k}$ to $t_{k+1}$ in (\ref{integralapproximation}) (namely $z=t_{k+1}$ is fixed) reduces to the Trapezoidal rule in numerical integration. If sampling-time intervals are uniform and we use Trapezoidal rule in (\ref{integralapproximation}), the continuous-time error considered in this paper is proportional to the usual discrete-time errors contemplated in machine learning; however, if the sampling-time intervals are not uniform, no matter we use Trapezoidal rule in (\ref{integralapproximation}) or not, continuous-time and discrete-time errors are different\footnote{Recall that for both uniform and non-uniform sampling, usual discrete-time errors in machine learning give the same quantity. However, continuous-time error gives different quantities for both cases, which depend on sampling-time intervals.}. \textit{In both uniform and non-uniform samplings, we show in this paper that we can develop incremental learning algorithms for ODE discovery by considering continuous-time error and using Trapezoidal rule in (\ref{integralapproximation}) (see Remarks 4 and 5 as well).} It should be noted that the accuracy of the proposed method depends on the accuracy of an integral approximation method that also depends on the lengths of sampling-time intervals.

\noindent
\textbf{Corollary 2:}
Consider Problem (\ref{minimizationformulation}). An optimal solution of (\ref{minimizationformulation}) can be approximated by polynomials of finite degree of at most $x_{1}^{N_{1}} \hdots x_{n}^{N_{n}}t^{N_{n+1}}$ from Algorithm \ref{algorithm1}.

\noindent
\textit{Proof.} See Appendix E.

As mentioned in \cite{Koopmandecember1}, existing results can also be divided into two categories: \textit{direct} methods and \textit{indirect} methods. Direct methods (like SINDy) assume that time derivatives of
the state can be accurately estimated, that may become prohibitive when the
sampling time is too low, the measurements too noisy, or the
time series too short (\cite{Koopmandecember1}). On the other hand, indirect methods (like NODEs) do not need the estimation
of time derivatives. This results in having less stringent requirements on the sampling rate, number of data points, and capability to handle irregularly spaced data points. As such our proposed method is an \textit{indirect} method.

\noindent
\textbf{Remark 5:}
An interesting aspect of Algorithm 1 is that once a datum is added to the data set, the matrices $\tilde{Q}_{k}$ and $P^{l}_{k}, l=1,\hdots,n,$ can be updated incrementally from $\tilde{Q}_{k-1}$ and $P^{l}_{k-1}, l=1,\hdots,n$ (see Appendix F). This property addresses the issues of memory restrictions, data security/privacy restrictions, and sustainable artificial intelligence (see \cite{tasksurvey333333333} for details). The incremental learning Algorithm 1 belongs to the category of \textit{Instance-Incremental Learning} where all training samples
belong to the same task and arrive one by one (see \cite{tasksurvey6666666} for details).

\noindent
\textbf{Remark 6:} It can be impractical to store or form the matrix $C$, defined in (\ref{matrixC}), when the dimension of the state (i.e., $n$) is large or $N_{1}, N_{2}, \hdots,N_{n+1}$ are large. Hence, we run into the \textit{curse of dimensionality}. This issue is analogous to those of other methods such as SINDy and EDMD, to name a few\footnote{SINDy Autoencoders (\cite{SINDyAutoencoders}) and residual DMD (\cite{ResidualDMD}), to name a few, have been proposed to overcome the curse of dimensionality of SINDy and EDMD, respectively.}.

\noindent
\textbf{Remark 7:} It is to be noted that the known external input $\mathcal{F}(t)$ in (\ref{1}) (if exists) can be \textit{any} functions; however, if there exists a hidden/unknown input, Algorithm 1 can exactly discover it if it is represented by polynomials of finite degree (see Examples 3 and 4).


\subsubsection{Sparsity-Promoting FINS (FINS\_{s})}

  Since for each dimension $l=1,\hdots,n,$ of $\dot{x}(t)$ we have $2^{(N_{1}+1)(N_{2}+1) \hdots (N_{n+1}+1)}-1$ choices (namely each parameter of the $l^{th}$ row of matrix $C$ defined in (\ref{matrixC}) exists or not), we have $(2^{(N_{1}+1)(N_{2}+1) \hdots (N_{n+1}+1)}-1)^{n}$ combinatorial structures of dynamical systems for the  vector field $\hat{f}(t,x)$, defined in Algorithm 1, to be learned. Due to the fact
that most physical systems have only a few relevant terms that
define the dynamics, the unknown vector field $f(t,x)$, if represented by polynomials, is sparse in the
space $\mathbb{R}^{n \times (N_{1}+1)(N_{2}+1) \hdots (N_{n+1}+1)}$. Hence, instead of solving (\ref{corollarylinearalgebraicequation}), we wish to seek solutions of the following optimization problem for each $l=1,\hdots,n$:

\begin{equation}\label{finsssssss11111}
	\begin{aligned}
		& \underset{C_{k}^{l}}{\text{min}}
		& & \Vert C_{k}^{l} \Vert_{0} \\
		& \text{s.t.}
		& & 2\tilde{Q}_{k} C^{l}_{k}=-P^{l}_{k}
	\end{aligned}
\end{equation}
in which $\Vert . \Vert_{0}$ represents $l_{0}$-norm (see Notations in Section 2). Problems (\ref{finsssssss11111}) are known as \textit{compressed sensing (or compressive sensing) or cardinality minimization} (\cite{comprseedsensensingsurvey1111111}, \cite{comprseedsensensingsurvey3333333}, \cite{cardinalityminimization}). Problems (\ref{finsssssss11111}) can also be expressed as  
\begin{equation}\label{finsssssss222222}
	\begin{aligned}
		& \underset{C_{k}^{l}}{\text{min}}
		& & \Vert C_{k}^{l} \Vert_{0} \\
		& \text{s.t.}
		& & \Vert 2\tilde{Q}_{k} C^{l}_{k}+P^{l}_{k} \Vert_{2} \leq \mu_{l} 
	\end{aligned}
\end{equation}
where $\mu_{l} \geq 0, \mu_{l} \in \mathbb{R}, l=1,\hdots,n,$ are tolerances. The $l_{0}$-norm is a non-convex, non-smooth, and integer valued function, and
the $l_{0}$-norm minimization problems are NP-hard problems that are generally not computationally tractable (see \cite{Compressedsensing1}, \cite{Compressedsensing2}). Several methods have been proposed to approximate a solution of (\ref{finsssssss11111}) and (\ref{finsssssss222222}) such as convex relaxation\footnote{One may think of imposing some well-known conditions such as  mutual coherence, the restricted isometry property (RIP), and the nullspace property (NSP) to guarantee that solutions of (\ref{finsssssss11111}) and (\ref{finsssssss222222}) are equal to the solutions of their convex relaxation problems. While evaluating the
mutual coherence of a given matrix is computationally easy, the
sparsity levels for which the mutual coherence can guarantee
recoverability are often too small to be of practical use (\cite{RIPdecemberRIP}). Moreover, evaluating RIP and NSP is in general NP-hard (see \cite{RIPdecemberRIP}). } (namely substituting $\Vert . \Vert_{1}$ for $\Vert . \Vert_{0}$ in (\ref{finsssssss11111}) and (\ref{finsssssss222222})) or iterative thresholding algorithms (\cite{Compressedsensing1}, \cite{Compressedsensing2}, \cite{comprseedsensensingsurvey1111111}, \cite{comprseedsensensingsurvey3333333}). We call \textit{any} algorithms to solve (\ref{finsssssss11111}) or (\ref{finsssssss222222}) \textit{FINS\_s}.

\subsubsection{Discussion on Non-Polynomial Vector Fields}

Mathieu’s (or Mathieu-Hill's) equation is one of the archetypical equations of nonlinear vibrations theory that is not only
associated with this field, but also it
appears in applied mathematics as well as many engineering fields (\cite{MathieuHillequation}, \cite{Mathiuequationdecember}). The classical\footnote{Generalization of Mathieu's equation has also been considered (see \cite{Mathiuequationdecember}). } Mathieu's equation is
\begin{equation}\label{mathieuformulation1}
    \frac{d^{2}x(t)}{dt^{2}}+(\delta+\epsilon  \cos(t))x(t)=0
\end{equation}
where $\delta \in \mathbb{R}$ and $\epsilon \in \mathbb{R}$ are constant. The state space representation of (\ref{mathieuformulation1}) by changing variables $x_{1}(t):=x(t)$ and $x_{2}(t):=\dot{x}(t)$ is
\begin{equation*}
    \begin{pmatrix}
			\dot{x}_{1}(t) \\
			\dot{x}_{2}(t) 
		\end{pmatrix}=\begin{pmatrix}
			x_{2}(t) \\
			-(\delta+\epsilon  \cos(t))x_{1}(t) 
   		\end{pmatrix}.
\end{equation*}
Hence, the vector field is $f(t,x_{1},x_{2}):=\begin{pmatrix}
			x_{2} \\
			-(\delta+\epsilon  \cos(t))x_{1} 
   		\end{pmatrix}$. It is easy to check that $\Vert f \Vert_{L^{2}_{\varphi}}$ (defined in (\ref{normf})) is bounded , i.e.,
        \begin{align*}
            \Vert f \Vert^{2}_{L^{2}_{\varphi}}&=\int_{\mathbb{R}^{3}} e^{-x_{1}^{2}-x_{2}^{2}-t^{2}} f^{T}(t,x_{1},x_{2}) f(t,x_{1},x_{2}) dx_{1} dx_{2} dt  < \infty
        \end{align*}      
        which implies that $f \in L^{2}_{\varphi}(\mathbb{R}^{3}, \mathbb{R}^{2})$. Therefore, the learning framework (\ref{minimizationformulation}) still \textit{includes} ODE discovery of the classical Mathieu's equation; however, as we mentioned earlier, due to computational restrictions, FINS (or FINS\_s) can approximate $f$ with polynomials. Although FINS and FINS\_s are not able to discover a vector field represented by non-polynomials completely (since such vector fields are not in the span of finite Hermite polynomials), they can discover those parts of the vector field which are (is) represented by polynomials if exist(s) and if the highest degrees of Hermite polynomials are greater than or equal to the corresponding degrees of the polynomial vector field (see Examples 6-7 for details).

\subsection{Computational Complexity of Algorithm 1}

\noindent
\textit{Time complexity of Algorithm 1:} Let define $\tau:=N_{1}N_{2} \hdots N_{n+1}.$ It is easy to compute that the time complexity of Step 1 in Algorithm 1 is $O(k \tau^{2}+kn \tau+kn+k)$ where $O$ denotes big o. The time complexity of Step 2 in Algorithm 1 is $O(n \tau^{3})$. Therefore, if a datum is added to $\mathcal{D}_{t}$ at each $t_{k}$, then the time complexity of Algorithm 1 will be $O(kn \tau^{5}+kn^{2} \tau^{4}+kn^{2} \tau^{3}+kn \tau^{3})$; while if the cardinality of $\mathcal{D}_{t}$ is fixed for the training time, then the time complexity of Algorithm 1 will be $O(n \tau^{3} +k \tau^{2}+kn \tau+kn+k)$.   

\noindent
\textit{Space complexity of Algorithm 1:} The largest matrix to be stored is $\tilde{Q}_{k}$. Hence, the space complexity is $O(\tau^{2}).$

The time complexity of Algorithm 1 increases linearly with respect to the number of training samples, i.e., $k$. The number of coefficients to be estimated in (\ref{888888888}) increases exponentially\footnote{Consider $N_{1}=N_{2}=\hdots=N_{n+1}=\tilde{N} \in \mathbb{N}$. Then $\tau=\tilde{N}^{n+1}$ which is exponential with respect to $n$.} with regards to the dimension (number of states) of the system, i.e., $n$. Although this is not a big concern for low dimensional systems, this increase of time complexity for high dimensional dynamical systems imposes a challenge that needs to be addressed in the future work (see also Remark 6).

To reduce the time complexity of solving (\ref{corollarylinearalgebraicequation}), we can decouple the computation among $\tilde{m} \geq 2, \tilde{m} \in \mathbb{N},$ processors where each processor only knows a subset of the rows of the partitioned matrix $[2 \tilde{Q}_{k},-P_{k}^{l}]$, and use a distributed algorithm over random networks, e.g., in \cite{alavianiLAE}. That makes the algorithm computationally tractable in the case of high-dimensional
systems. Precisely, it is easy to calculate that the time complexity of the algorithm in \cite{alavianiLAE} is $O((\underset{i=1,\hdots,\tilde{m}}{max} \{\mu_{i} \tau^{2} \}+\tau^{2}+\tilde{m} \tau) \tilde{k}n)$ where $\tilde{k}$ is the iteration of the algorithm in (\cite{alavianiLAE}), and  $\sum_{i=1}^{\tilde{m}} \mu_{i}=(N_{1}+1)(N_{2}+1) \hdots (N_{n+1}+1), \mu_{i} \in \mathbb{N}$; indeed, if $\mu_{i}=1, i=1, \hdots, \tilde{m},$ then it reduces to $O(n \tau^{2}).$ Also space complexity of the algorithm in \cite{alavianiLAE} is $O(\tau^{2})$. Consequently, if we use the distributed algorithm for solving the linear algebraic equation (\ref{corollarylinearalgebraicequation}) and if a datum is added to $\mathcal{D}_{t}$ at each $t_{k}$, then the time complexity of Algorithm 1 will be $O((k \tau^{2}+kn \tau+kn+k)(\underset{i=1,\hdots,\tilde{m}}{max} \{\mu_{i} \tau^{2} \}+\tau^{2}+\tilde{m} \tau) \tilde{k}n)$. If we use the distributed algorithm for solving the linear algebraic equation (\ref{corollarylinearalgebraicequation}) and if the cardinality of $\mathcal{D}_{t}$ is fixed for the training time, then the time complexity of Algorithm 1 will be $O(k \tau^{2}+kn \tau+kn+k+(\underset{i=1,\hdots,\tilde{m}}{max} \{\mu_{i} \tau^{2} \}+\tau^{2}+\tilde{m} \tau) \tilde{k})m$.   

\noindent
\textit{Computational time of Algorithm 1:} Algorithm 1 is very efficient for the learning process practically. Its computational time is comparable to SINDy's, while, unlike SINDy, Algorithm 1 is an incremental learning algorithm.  We show the evaluation of computational time with Duffing, Lorenz, and Lorenz 96 systems in Subsection 4.1 below. 
\begin{table*}[t!]
  \centering    
  
  \scriptsize
  \begin{tabular}{llll}
  Dynamical systems & Equations & Parameters\\
  \hline
     Duffing equation (Unforced) & 
		$\begin{pmatrix}
			\dot{x} \\
			\dot{y} 
		\end{pmatrix}=\begin{pmatrix}
			y \\
			-\delta y-\alpha x-\beta x^{3} 
   		\end{pmatrix}$ & $\delta=\alpha=\beta=1$ \\
\hline
    Van der Pol & $\begin{pmatrix}
			\dot{x} \\
			\dot{y} 
   		\end{pmatrix}=\begin{pmatrix}
			y \\
			-\alpha x+\varepsilon_{1}y-\varepsilon_{2} x^{2}y 
   		\end{pmatrix}$& $\alpha=1, \varepsilon_1 = \varepsilon_2= 0.05$ \\
\hline
    Duffing equation with known force & $\begin{pmatrix}
			\dot{x} \\
			\dot{y} 
   		\end{pmatrix}=\begin{pmatrix}
			y \\
			-\delta y-\alpha x-\beta x^{3} 
   		\end{pmatrix}+\begin{pmatrix}
		    0\\
      1000\sin(2t)
		\end{pmatrix}$ & $\delta=\alpha=\beta=1$\\
 \hline
     Van der Pol with unknown force & $\begin{pmatrix}
			\dot{x} \\
			\dot{y} 
   		\end{pmatrix}=\begin{pmatrix}
			y \\
			-\alpha x+\varepsilon_{1}y-\varepsilon_{2} x^{2}y+\delta t^{2} 
   		\end{pmatrix}$  & $\alpha=1, \varepsilon_1 = \varepsilon_2= 0.02, \delta = 0.01$ \\
\hline
 Lorenz chaotic system & $
		\begin{pmatrix}
			\dot{x} \\
			\dot{y} \\
   \dot{z}
		\end{pmatrix}=\begin{pmatrix}
			10(y-x) \\
			x(28-z)-y \\
   x y-\frac{8}{3}z
		\end{pmatrix}
	$ & $\alpha= 10, \rho= 28, \beta=8/3$\\
     \hline
Mathieu-Hill's equation & $\begin{pmatrix}
			\dot{x} \\
			\dot{y} 
		\end{pmatrix}=\begin{pmatrix}
			y \\
			-(\delta+\epsilon  \cos(t))x 
   		\end{pmatrix}$ & $\delta=1, \epsilon=0.2$ \\
     \hline
 & &\\
  \end{tabular}
\caption{Dynamical systems considered in Examples 1-7}

  \label{table1}
\end{table*}

\section{Demonstration on Numerical Examples}

We test Algorithm 1 on several numerical examples,  as shown in Table \ref{table1}. \textit{We use an improved accuracy for differentiation by averaging one step forward and one step backward for SINDy in all examples}. To fairly compare with SINDy, we use the \textit{sequential
threshold least-squares} for FINS\_s and select the threshold value of sparsity to be 0.025 in all examples unless mentioned. 
Note that ``np" and ``sp" in all examples denote non-sparse and sparse, respectively. We should mention that to fairly compare with existing methods, we consider, in all tables, $ Error := \frac{1}{p} \sum_{i=1}^{p} \Vert x_{i}-\hat{x}_{i} \Vert_{2}$ where $x_{i}$ and $\hat{x}_{i}$ are the true state and estimated state, respectively, and $p$ is the number of points used for error calculation. Note also that {\tt ode45} solver in MATLAB R2024b is used to create our datasets by solving the systems of differential
	equations. We mention that we use all data for training in all methods (see Subsection 3.1).

    \begin{figure}[t!]
   \begin{minipage}{0.48\textwidth}
     \centering
     \includegraphics[width=0.95\linewidth]{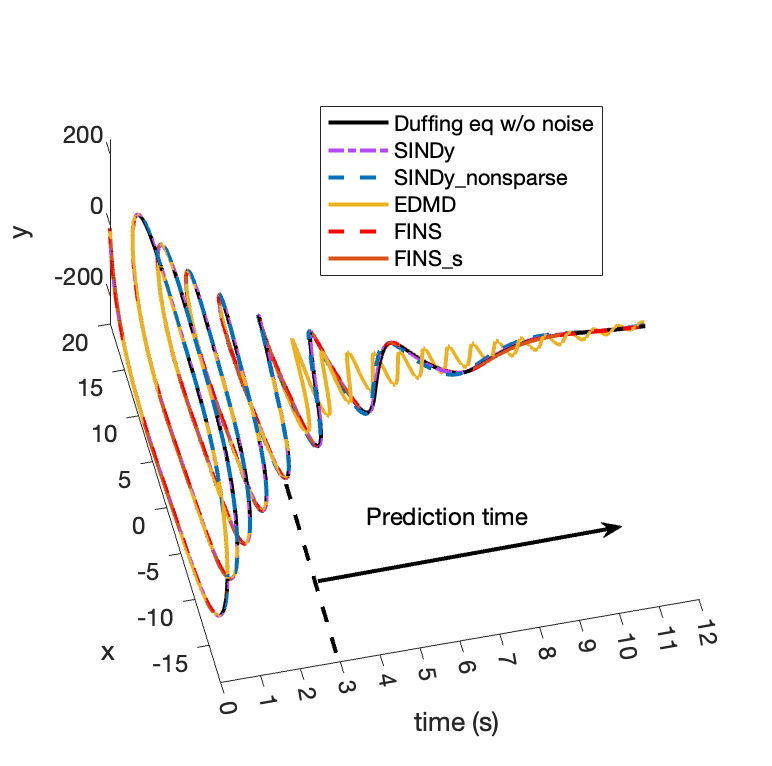}
     \caption{3D plot of trajectories of learned models by several methods for Duffing equation without noise in Example 1.}\label{figure11}
   \end{minipage}\hfill
\begin{minipage}{0.48\textwidth}
\centering
   \includegraphics[width=0.96\linewidth]{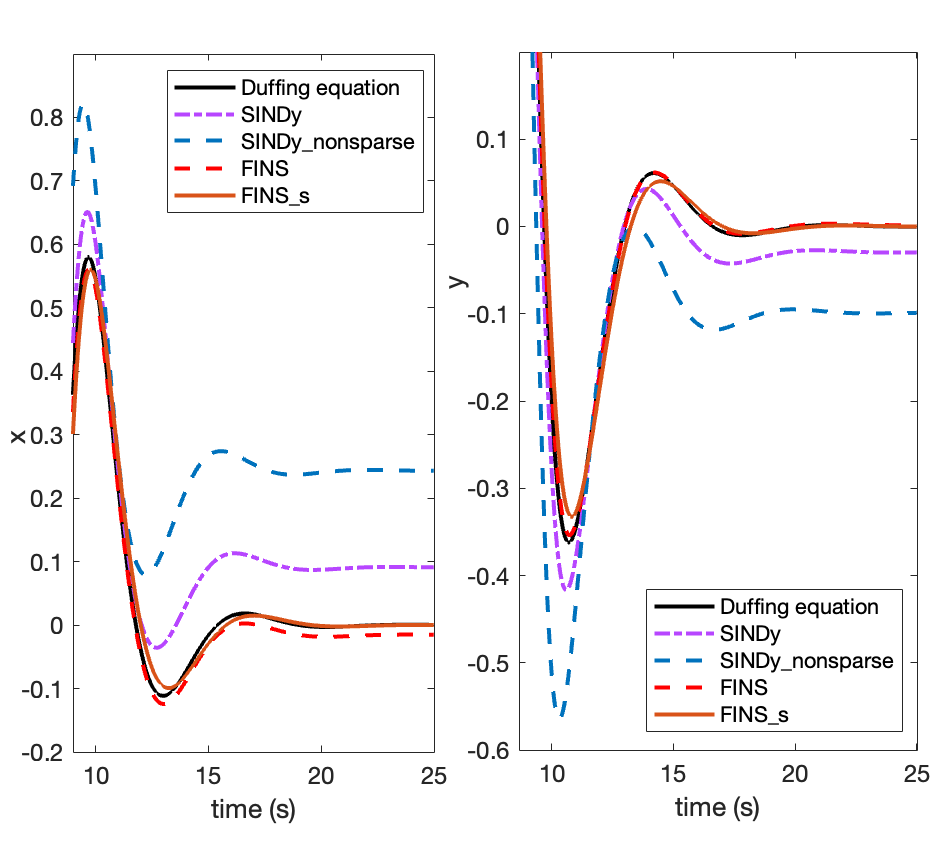}
		\caption{Zoomed Plots of Variables $x(t)$ and $y(t)$ of Duffing equation in Example 1 without noise.}
	\label{figure33333333}
   \end{minipage}
\end{figure}

\subsection{Evaluation of the Proposed Method Accuracy}

\noindent \textbf {Example 1: Discovering Duffing Equation}\\
We select training time $[0, 3]$ seconds with initial conditions $x(0)=20$ and $y(0)=-10$ to simulate the dataset. The {\tt ode45} solver gives us a data set $\mathcal{D}_{3}$ with cardinality equal to 1136 with \textit{non-uniform} sampling time. Consequently, $k(3)=1135$. We select the highest degree of Hermite polynomials\footnote{Note that we did not augment $x$ and $y$ with $t$ to capture the nonlinearity in time-invariant systems.} for both variables $x$ and $y$ to be 3, namely $N_{1}=N_{2}=3$. 

We select $w(t) \equiv 1, t \in [0,3]$ (see Remark 1) for simulation and apply Algorithm 1 to obtain the corresponding nonlinear functions at $k=1135$ with 0.0001 accuracy for FINS and FINS\_s. We also simulated SINDy\footnote{gEDMD (\cite{gEDMDDD}) reduces to SINDy in all numerical examples since deterministic dynamical systems are considered.} with the highest monomial of degree 3, i.e., $x^{3},y^{3}, xy^2,yx^2$. Since SINDy does not include monomial terms having degree of higher than 3, like $x^2y^2, x^2y^3, x^3y^3$, etc., the SINDy library is only using 10 terms while FINS (and FINS\_s) is using 16 terms.  This gives SINDy an advantage of a smaller search space including all true terms. We use Hermite polynomials of degree 3 as lifting functions in EDMD\footnote{We are aware that EDMD (\cite{wiliams}) works under uniform sampling time (see below where we also compared the methods under \textit{uniform} sampling time), but we show that it is not promising for non-uniform sampling time although the maximum sampling-time interval in the dataset is $0.0071$ second.}. We obtain the learned vector field from SINDy as
\begin{align*}
    \dot{x}&=0.9998y+0.0294\\
	\dot{y}&=0.0674-1.0585x-0.9994y-0.9995x^{3}
\end{align*}
from FINS as  
			\begin{align*}		\dot{x}&=-0.0015+1.0004y-0.0007x \\
	\dot{y}&=-0.0140-1.0004y-1.0070x+0.0001xy+0.0002x^{2}-x^{3}
   		\end{align*}   
and from FINS\_s  as 
\begin{align*}
    \dot{x}&=1.0001 y \\
	\dot{y}&=-1.0006y-0.9338x-1.0003x^{3}.
\end{align*}

The learned results are plotted in Figures \ref{figure11} and \ref{figure33333333}. We simulated the learned dynamics of SINDy and EDMD for prediction time $[3,25]$ seconds in the figures\footnote{Due to the poor extrapolation capability of EDMD, the result of EDMD shows high oscillatory (as can be seen in Figure 3) and is omitted in Figure 4.} that show the capability of the proposed method to learn and predict the underlying nonlinear dynamics using training data \textit{without noise}. The error of the methods for prediction time $[3, 25]$ are provided in Table 2. It can be seen that the prediction error of FINS is approximately 1 order smaller than that of SINDy, and 2 orders smaller than that of EDMD for Duffing equation without noise. It is shown that the nonlinearity is learned with high accuracy by the corresponding nonlinear terms without involving $t$.  \textit{Our simulations imply that by reducing the sampling-time intervals, increasing the training time, and/or reducing the degree of Hermite polynomials (but not less than 1 for $y$ and 3 for $x$), the exact unknown dynamics can be recovered.}

\begin{table}[t!]

\begin{center}
	\begin{tabular}{|c|c|c|c|}
		\hline
		\textbf{\textbf{\vtop{\hbox{\strut Duffing}\hbox{\strut equation}}}}& \textbf{EDMD} &  \textbf{SINDy}& \textbf{FINS}\\
		\hline
		{no noise}  & 13.72 $\pm$ 11.30 &  {\vtop{\hbox{\strut 0.48 $\pm$ 0.68 (np) }\hbox{\strut 0.09 $\pm$ 0.03 (sp)}}} & {\vtop{\hbox{\strut \textbf{0.03 $\pm$ 0.06 }(np) }\hbox{\strut \textbf{0.05 $\pm$ 0.09} (sp)}}} \\
		\hline
		{noisy}  & 8.87 $\pm$ 5.34 &  {\vtop{\hbox{\strut 4.88 $\pm$ 3.84 (np) }\hbox{\strut 6.36 $\pm$ 5.02 (sp)}}}& {\vtop{\hbox{\strut \textbf{2.91 $\pm$ 2.11} (np) }\hbox{\strut 5.20 $\pm$ 3.97 (sp)}}} \\
		\hline
	\end{tabular}
	\label{tttable1119}
 \caption{Comparison of errors of prediction time (i.e., $[3,25]$ and $[6,10]$ seconds for no noise and noisy cases, respectively) for learned models of different methods for Duffing system in Example 1 under \textit{non-uniform} sampling time}
\end{center}
\label{table1example1error2}
\end{table}


To demonstrate the robustness to real measured data with noise\footnote{By \textit{robustness to noise} we mean how much the results under noisy data deviate from the results without noisy data.}, we added an independent and identically distributed (i.i.d.) Gaussian noise with a level corresponding to a signal-to-noise ratio (SNR) to be equal to 20 with regards to the root-mean-square value of the signal for the training time $[0,6]$ with $k(6)=1549$ under \textit{non-uniform} sampling time. We select the highest degree of Hermite polynomials of both variables $x$ and $y$ to be 3, namely $N_{1}=N_{2}=3$. FINS yields the following dynamics at $k=1549$:
\begin{align*}
\dot{x}&=0.1699+0.9884 y+0.0943 x-0.0009 x y + 0.0055 x^{2}+0.0002x^2y-0.0013 x^{3} \\
\dot{y}&=-1.6766-0.9501 y-0.7264 x -0.9934 x^{3} +0.0416 x^{2} + 0.0015 y^{2} +0.0353 x y \\
&\quad{}+0.0005 y x^{2}-0.0003x^3y
\end{align*}
and FINS\_s provides
\begin{align*}
    \dot{x}&=0.1601+0.9941y+0.0740x \\
    \dot{y}&=2.5819-0.9427y-59.0495x-0.7265 x^{3}
\end{align*}
while SINDy yields
\begin{align*}
\dot{x}&=0.1180+1.0021 y+0.2380 x, \\
\dot{y}&=9.0973-0.9750 y-1.8133 x -0.9951 x^{3}-0.0996 x^{2}
\end{align*}

\begin{figure}[t!]
   \begin{minipage}{0.48\textwidth}
     \centering
     \includegraphics[width=1.06\linewidth]{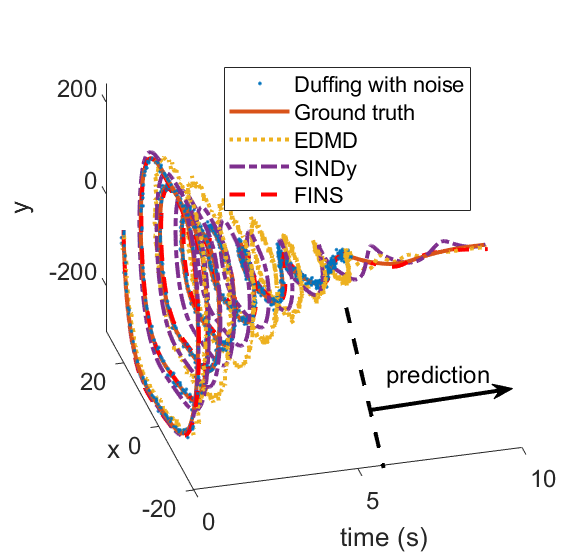}
     \caption{3D plot of trajectories of learned models by several methods for Duffing equation in Example 1 with additional Guassian noise (SNR = 20)}\label{figure3331}
   \end{minipage}\hfill     
   \begin{minipage}{0.5\textwidth}
     \centering
     \includegraphics[width=0.82\linewidth]{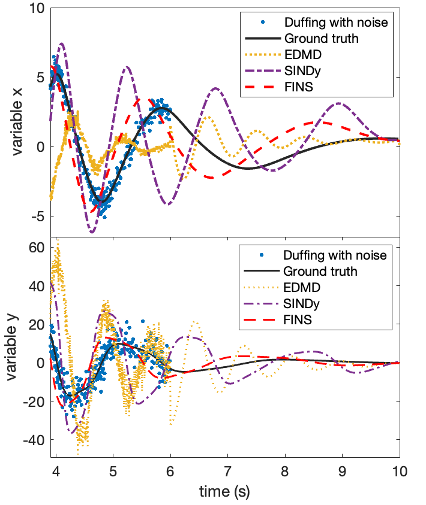}
     \caption{Variables $x(t)$ and $y(t)$ of learned models of several methods for Duffing equation with additional Gaussian noise (SNR = 20) in Example 1}\label{figure3332}
   \end{minipage}
\end{figure}

Note that the learned dynamics may vary with a different realization of Guassian noise. The results given by Algorithm \ref{algorithm1} with noisy data for training and prediction time are shown in Figures \ref{figure3331}-\ref{figure3332}. The errors of the methods for prediction time, i.e., $[6,10]$ seconds, are given in Table 2. The results in Figures \ref{figure11}-\ref{figure3332} and Table 2 validate that Algorithm \ref{algorithm1} can learn the nonlinear dynamics and clearly present its robustness for noisy data. However, it is worth to mention that even if FINS\_{s} often yields sparser results than FINS, both FINS\_{s} and SINDy appear to be less robust compared to FINS for noisy data. While incorporating sparsity constraints into the proposed algorithm leads to more compact representations, it does so at the expense of reduced robustness—a trade off that is also observed in SINDy, as mentioned in (\cite{sindy}). This opens a subject for further research to develop a robust FINS\_s (see Subsection 3.2.1). It should also be noted that our simulations show that in the case of noisy data, both SINDy and FINS\_s are highly dependent on the choice of the threshold value of sparsity (which in this example is 0.025). 

Next, we fairly compare the proposed method with other methods which require \textit{uniform} sampling time such as Koopman-based lifting technique (KLT) proposed in \cite{Koopmandecember1}. For uniform sampling time for Duffing equation without noise, we consider the training time $[0,1.15]$ seconds under sampling time $0.001$ seconds (we also try different unifrom sampling time in Example 2). Now we rewrite the Duffing equation in Table 1 as 
$$\begin{pmatrix}
			\dot{x} \\
			\dot{y} 		\end{pmatrix}=\begin{pmatrix}
			constant1+\gamma y + \eta x \\
			constant2-\delta y-\alpha x-\beta x^{3} 
   		\end{pmatrix}$$

We select $m=m_{F}=3$ for KLT that, similar to SINDy, uses 10 monomial terms, while FINS (and FINS\_s) uses 16 terms; this gives KLT and SINDy an advantage of a smaller search space including all true terms. The learned coefficients by SINDy, KLT, FINS, and FINS\_s are given in Table 3. Since the coefficients are close, we select longer times for prediction, i.e., $[1.15,500]$ seconds, to calculate prediction errors. We also computed the prediction errors of EDMD, and the results are provided in Table 4.\\

\begin{table}[t!]
	
	\begin{center}
		\begin{tabular}{|c|c|c|c|c|c|c|c|}
			\hline
			Parameters & $\delta$ & {\textbf{$\alpha$}} & {\textbf{$\beta$}} & {\textbf{$\gamma$}} & {\textbf{$\eta$}} & {constant1} & {constant2}\\
			\hline
			Duffing equation  & 1 & 1 & 1 & 1 & 0 & 0 & 0\\
			\hline
			SINDy & 1.0051 & 0.7393 & 1.0007 & 0.9995 & 0 & -0.1005 & -1.2393\\
			\hline
			KLT  & 0.9999 & 1.0002 & 0.9999 & 1.0000 & 0 & 0 & 0.0005\\
			\hline
			FINS & 1.0004 & 0.9709 & 1.0003 & 1.0000 & -0.0006 & -0.0002 & -0.0077 \\
			\hline
			FINS\_s  & 1.0001 & 0.9991 & 1.0000 & 1.0000 & 0 & 0 & 0\\			
			\hline			
		\end{tabular}
		\label{table1113AppG}
\caption{Learned models by several methods in Example 1 without noise under \textit{uniform} sampling time}
	\end{center}
\end{table}

\begin{table}[t!]
\begin{center}
	\begin{tabular}{|c|c|c|c|c|}
		\hline
		\textbf{\textbf{\vtop{\hbox{\strut Duffing}\hbox{\strut equation}}}}& \textbf{EDMD} & \textbf{KLT}& \textbf{SINDy}& \textbf{FINS}\\
		\hline
		{no noise}  &  {\vtop{\hbox{\strut \text{1.5141$\times 10^{99} \pm$ } }\hbox{\strut \text{1.7720$\times 10^{100}$}}}} & 0.0006$\pm$0.0020 & {\vtop {\hbox{\strut 0.8988$\pm$0.1949}}} & {\vtop{\hbox{\strut \text{0.0113$\pm$0.0326 }(np) }\hbox{\strut \textbf{0.0003$\pm$0.0040} (sp)}}} \\
		\hline
		\end{tabular}
	\label{tttable1119}
 \caption{Comparison of errors of learned models of different methods for Duffing system in Example 1 under \textit{uniform} sampling time}
\end{center}
\label{table1example1error2AppG}
\end{table}

\noindent
\textbf{Example 2: Discovering Van der Pol Equation} \\
We first rewrite the Van der Pol equation given in Table \ref{table1} with learnable coefficients as follows:
\begin{equation}\label{exmaple1stateequ11111}
	\begin{pmatrix}
		\dot{x} \\
		\dot{y} 
	\end{pmatrix}=\begin{pmatrix}
		const1 + \beta y\\
		
		const2 - \alpha x + \varepsilon_1 y -\varepsilon_2 x^{2}y
	\end{pmatrix}.
\end{equation}
To learn the Van der Pol equation, we obtain $k(20)=1545$ for the training time $[0,20]$ seconds with \textit{non-unifrom} sampling time and initial conditions $x(0)=10$ and $y(0)=-5$. We select the highest degree of Hermite polynomials of both variables $x$ and $y$ to be 3, namely $N_{1}=N_{2}=3$. We apply Algorithm 1 to obtain the nonlinear function at $k=1545$ as
\begin{align*}
	\dot{x}&=0.0001+1.0001 y\\
	\dot{y}&=- x +0.0500 y-0.0500 x^{2} y
\end{align*}

\begin{figure}[t!]
 \begin{center}
 \centerline{\includegraphics[width=0.7\linewidth]{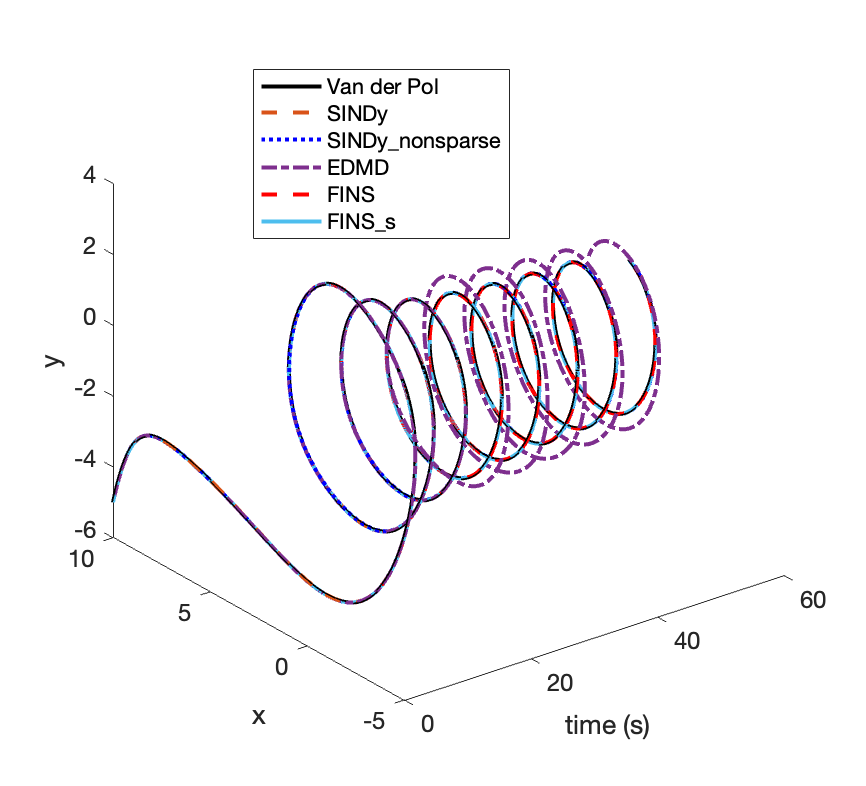}}
     \caption{3D plot of trajectories of learned models by several methods for Van der Pol equation in Example 2}
     \label{figure31}
\end{center}
\end{figure}

\begin{table}[t!]
	
	\begin{center}
		\begin{tabular}{|c|c|c|c|c|c|c|}
			\hline
			Parameters & $\beta$ & {\textbf{$\alpha$}} & {\textbf{$\varepsilon_1$}} & {\textbf{$\varepsilon_2$}} & {const1} & {const2}\\
			\hline
			Van der Pol  & 1 & 1 & 0.05 & 0.05 & 0 & 0\\
			\hline
			SINDy & 0.9999 & 0.9991 & 0.0500 & 0.0499 & 0 & 0\\
			\hline
			SINDy\_np  & 0.9995 & 0.9983 & 0.0517 & 0.0501 & -0.0015 & 0.0068\\
			\hline
			FINS & 1.0001 & 1 & 0.0500 & 0.0500 & 0.0001 & 0 \\
			\hline
			FINS\_s  & 0.9999 & 1 & 0.0500 & 0.0500 & 0 & 0\\
			
			\hline
			
		\end{tabular}
		\label{table1113}
  \caption{Learned models by SINDy and FINS in Example 2 under \textit{non-uniform} sampling time}
	\end{center}
\end{table}

\begin{table}[t!]
	
	\begin{center}
		\begin{tabular}{|c|c|c|c|}
			\hline
			& \textbf{EDMD} &  \textbf{SINDy}&  \textbf{FINS}\\
			\hline
			\textbf{\vtop{\hbox{\strut Van Der}\hbox{\strut Pol}}}  & 4.0798$\pm$  1.5080 &  {\vtop{\hbox{\strut 0.0829 $\pm$ 0.0224 (np) }\hbox{\strut 0.0345 $\pm$ 0.0110 (sp)}}} & {\vtop{\hbox{\strut \textbf{0.0044 $\pm$ 0.0010 (np)} }\hbox{\strut 0.0046 $\pm$ 0.0010 (sp)}}} \\
			\hline

		\end{tabular}
  \caption{Comparison of errors of learned models of different methods in Example 2 under \textit{non-uniform} sampling time}
		\label{table1112}
	\end{center}
\end{table}



\begin{table}[t!]
	
	\begin{center}
		\begin{tabular}{|c|c|c|c|c|c|c|}
			\hline
			Parameters & $\beta$ & {\textbf{$\alpha$}} & {\textbf{$\varepsilon_1$}} & {\textbf{$\varepsilon_2$}} & {const1} & {const2}\\
			\hline
			Van der Pol  & 1 & 1 & 0.05 & 0.05 & 0 & 0\\
			\hline
			SINDy & 0.9994 & 0.9994 & 0.0494 & 0.0499 & 0 & 0\\
			\hline
			KLT  & 1.0000 & 0.9999 & 0.0499 & 0.0499 & 0 & 0\\
			\hline
			FINS & 1.0000 & 1.0001 & 0.0500 & 0.0500 & 0 & 0 \\
			\hline
			FINS\_s  & 0.9999 & 1.0000 & 0.0499 & 0.0499 & 0 & 0\\
			
			\hline
			
		\end{tabular}
		\label{table1113AppH}
  \caption{Learned models by several methods in Example 2 under \textit{uniform} sampling time}
	\end{center}
\end{table}

\begin{table}[t!]

\begin{center}
	\begin{tabular}{|c|c|c|c|c|}
		\hline
		& \textbf{EDMD} & \textbf{KLT}& \textbf{SINDy}& \textbf{FINS}\\
		\hline
		 \textbf{\textbf{\vtop{\hbox{\strut Van der}\hbox{\strut Pol}}}} &  0.7817$\pm$0.4096  & 0.0068$\pm$0.0017 & {\vtop {\hbox{\strut 0.0337$\pm$0.0105}}} & {\vtop{\hbox{\strut \textbf{0.0038$\pm$0.0009 }(np) }\hbox{\strut \text{0.0067$\pm$0.0016} (sp)}}} \\
		\hline
		\end{tabular}
	\label{tttable1119}
 \caption{Comparison of errors of learned models of different methods for Van der Pol system in Example 2 under \textit{uniform} sampling time}
\end{center}
\label{table1example1error2AppH}
\end{table}

We also simulated SINDy with the highest monomials of degree 3, i.e., up to $x^{3}$ and $y^{3}$, as well as by using Hermite polynomials of at most degree 3 as lifting functions in EDMD (see also footnote 23). The simulation results are shown in Figure \ref{figure31}, and learned coefficients are shown in Table 5. The errors of the methods for prediction time $[20,60]$ seconds are provided in Table 6. 

Next we fairly compare the proposed method with other methods which require \textit{uniform} sampling time such as KLT. We consider the training time $[0,15]$ seconds under uniform sampling time having length $0.01$ seconds. We select $m=m_{F}=3$ for KLT that, similar to SINDy, uses 10 monomial terms, while FINS (and FINS\_s) uses 16 terms, that gives both KLT and SINDy an advantage of a smaller search space including all true terms. The learned coefficients by SINDy, KLT, FINS, and FINS\_s are given in Table 7\footnote{We chose the threshold value of sparsity to be 0.005 in Example 2.}. We consider a prediction time, i.e., $[15,50]$ seconds, for computing prediction errors, and the results are presented in Table 8.\\

\begin{figure}[t!]
   \begin{minipage}{0.48\textwidth}
     \centering
     \includegraphics[width=1.01\linewidth]{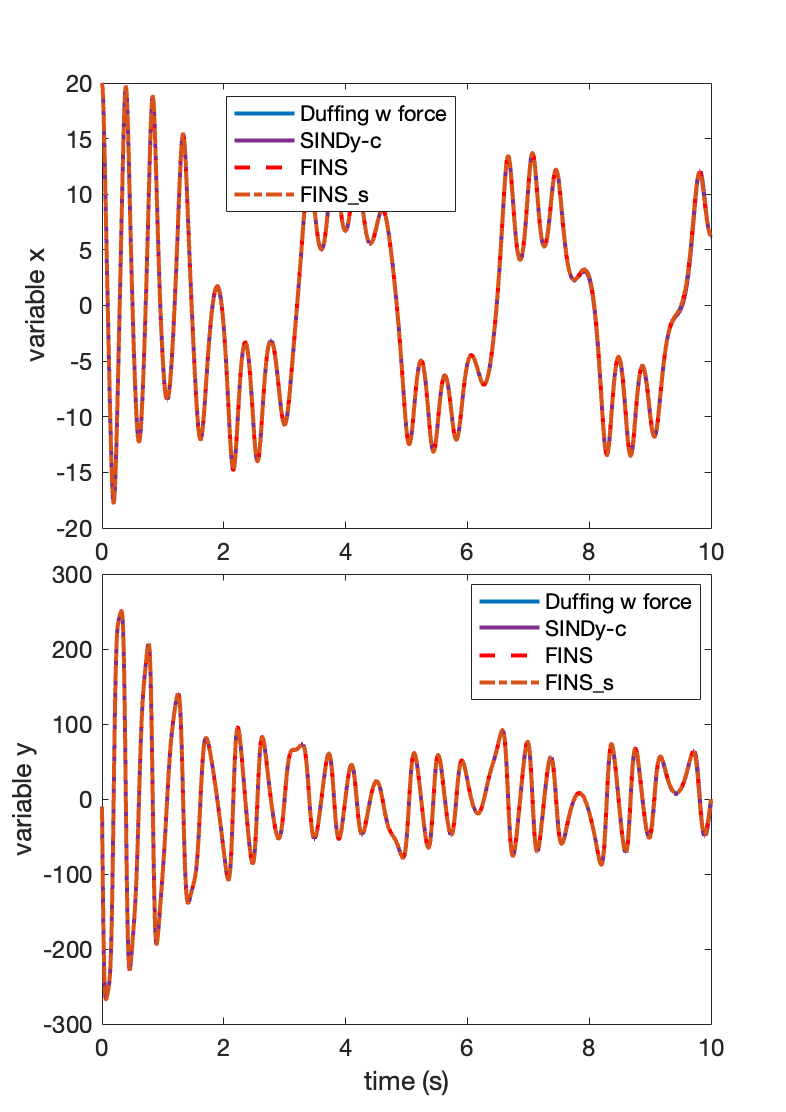}
     \caption{Variables $x(t)$ and $y(t)$ of learned models by SINDy\_c, FINS, and FINS\_s in Example 3}\label{figure1force0}
   \end{minipage}\hfill
   \begin{minipage}{0.48\textwidth}
     \centering
     \includegraphics[width=1\linewidth]{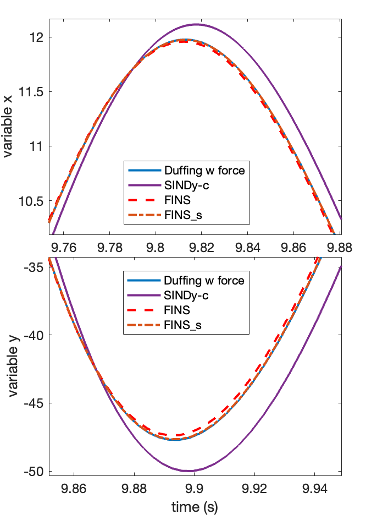}
     \caption{Zoomed Figure \ref{figure1force0} for better comparison}\label{figure2force1}
   \end{minipage}
\end{figure}

\noindent
\textbf{Example 3: Discovering Duffing Equation with Known External Force} \\
The Duffing equation with a known external force considered in this example is 	\begin{equation}\label{dufff1111111}
	\ddot{x}+ \delta \dot{x}+\alpha x+ \beta x^{3}=1000\sin(2t)
\end{equation}
where $\delta=\alpha=\beta=1$ and $\mathcal{F}(t):=1000\sin(2t)$. We select training time $[0, 5]$ seconds with initial conditions $x(0)=20$ and $y(0)=-10$ to make the dataset. The {\tt ode45} solver gives us a data set $\mathcal{D}_{5}$ having cardinality equal to 1772 with \textit{non-uniform} sampling time. Consequently, $k(5)=1771$. Similar to Example 1, we select $N_{1}=N_{2}=3$ and $w(t) \equiv 1, t \in [0,5]$, for the simulation. Then, we apply Algorithm 1 to obtain the following nonlinear functions at $k=1771$ with 0.0001 accuracy for FINS as
\begin{align*}
	\dot{x}&=-0.0026+1.0002 y+0.0005 x \\
	\dot{y} &=-0.1807-0.9980y -0.9635x  +0.0019x^{2}+0.0002xy-1.0003x^{3}+1000\sin(2t) 
\end{align*}
and for FINS\_s as
\begin{align*}
	\dot{x}&=y \\
	\dot{y} &=-0.0290-1.0002y-1.0157x-0.9997x^{3}+1000\sin(2t) 
\end{align*}

\begin{figure}[t!]
	\begin{center}
		\centerline{\includegraphics[scale = 0.5]{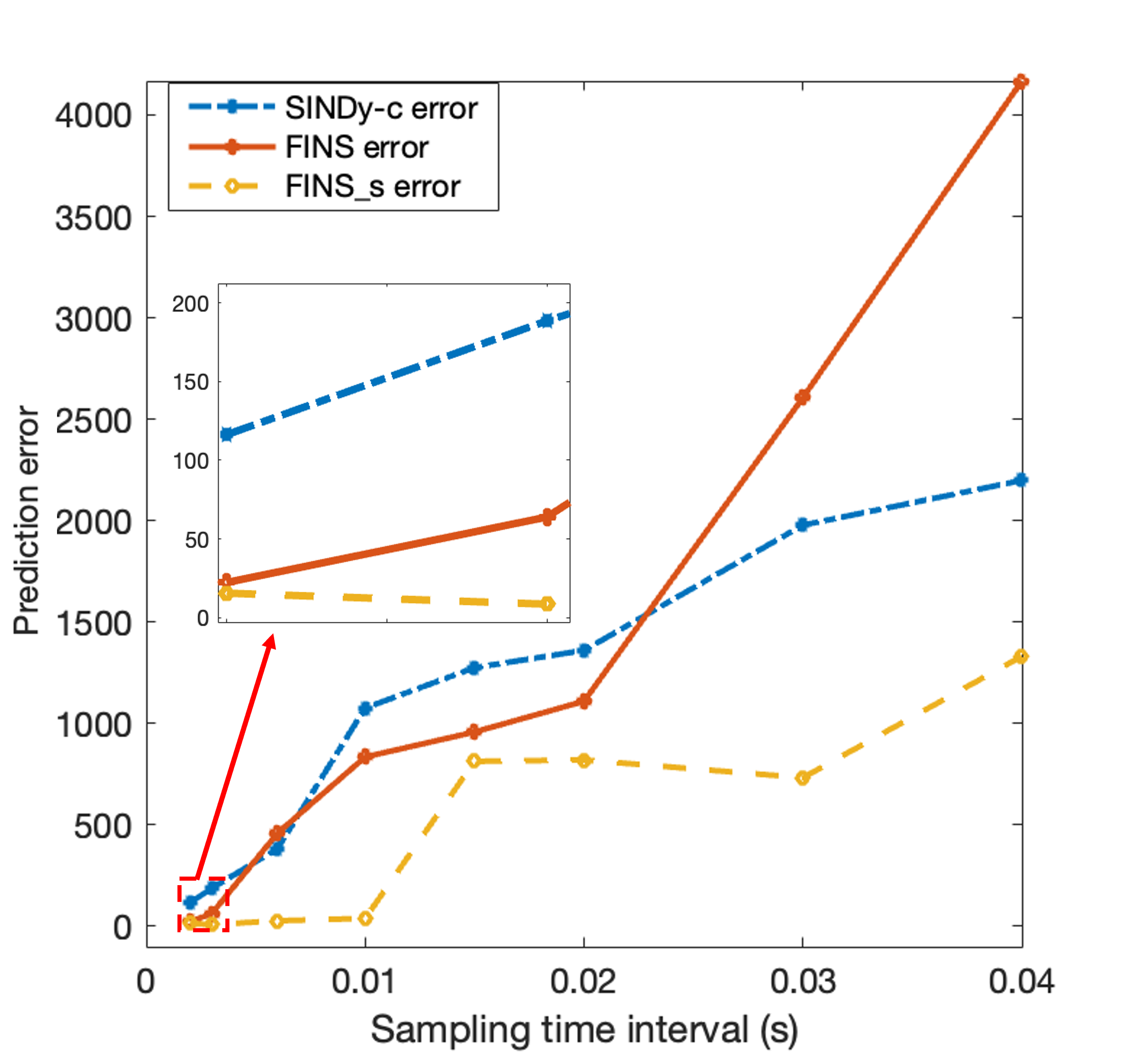}}
		\caption{Prediction error v.s. sampling time interval of learned models by SINDY\_c, FINS, and FINS\_s in Example 3}
		\label{figure2force2}
	\end{center}
\end{figure}

 The learning results are shown in Figures \ref{figure1force0}-\ref{figure2force1} where we also simulate the learned functions for prediction time $[5,10]$ seconds. To investigate the effect of sampling rate, we conducted evaluations using training data with different \textit{uniform} sampling times ranging from 0.002 to 0.04 second. The results are summarized and plotted in Fig. \ref{figure2force2} below in which we compared our method with SINDYc (namely SINDy with control input, see Subsection 1.1). We simulated SINDYc which provides the following dynamics:
\begin{align*}
	\dot{x}&=0.028+0.9997y \\
	\dot{y} &=-0.6149-0.9530x-0.9970y-0.9997x^{3}+1000\sin(2t) 
\end{align*}
Figures \ref{figure1force0}-\ref{figure2force2} show the capability of the proposed method to learn and predict the underlying nonlinear dynamics with a known external force.\\

\begin{figure}[t!]
   \begin{minipage}{0.48\textwidth}
     \centering
    \includegraphics[width=0.98\linewidth]{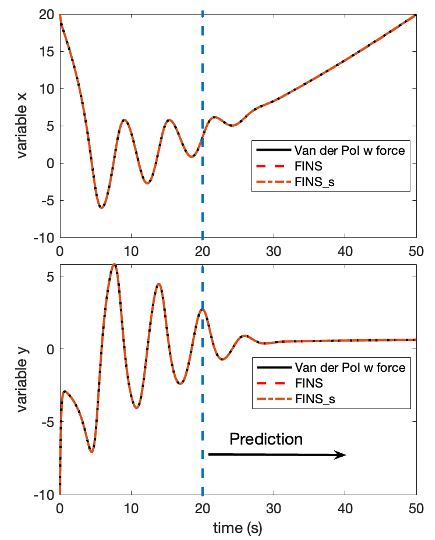}
     \caption{Learned $x(t)$ and $y(t)$ of by FINS and FINS\_s for Van der Pol system  with hidden external force}\label{figure5}
   \end{minipage} \hfill
\begin{minipage}{0.48\textwidth}
     \centering
     \includegraphics[width=0.9\linewidth]{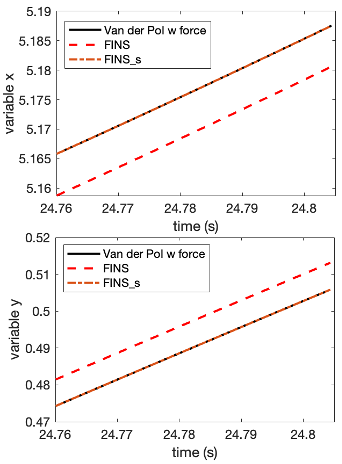}
     \caption{Zoomed Plot for Figure \ref{figure5} for Van der Pol system  with hidden external force}\label{figgg122222}
   \end{minipage}
\end{figure}



\noindent \textbf{Example 4: Discovering Van der Pol Equation Simultaneously with Hidden External Input} \\
Consider the following Van der Pol system in which the term $\delta t^{2} = 0.01 t^2$ can be viewed as an \textit{unknown} external input:
\begin{equation}\label{exmaple1stateequ11111}
	\ddot{x}+\alpha x-\varepsilon_1 \dot{x}  + \varepsilon_2 x^{2} \dot{x} = \delta t^2
\end{equation}

We obtain $k(20)=7888$ for training time $[0,20]$ seconds with \textit{non-unifrom} sampling time and initial conditions $x(0)=20$ and $y(0)=-10$. We select the highest degree of Hermite polynomials of variables $x,y,$ and $t$ to be 2, namely $N_{1}=N_{2}=N_{3}=2$. We apply Algorithm 1 to obtain the following nonlinear function at $k(20)=7888$ 
\begin{equation}\label{kkhh222222222222}
	\begin{pmatrix}
		
		\dot{x} \\
		\dot{y} 
		
	\end{pmatrix}=\begin{pmatrix}
		0.0008 + y -0.0001x- 0.0002t \\
		-0.0007 - 1.0001x + 0.0201y - 0.02x^2y + 0.0099t^2
	\end{pmatrix}
\end{equation}
which is under accuracy of 0.0001. With FINS\_s (with threshold value of sparsity to be 0.002), we can get the exact solution as:
\begin{equation}\label{kkhh222222222222}
	\begin{pmatrix}
		
		\dot{x} \\
		\dot{y} 
		
	\end{pmatrix}=\begin{pmatrix}
		0.9999y  \\
		-x + 0.0199y - 0.02x^2y + 0.01t^2
	\end{pmatrix}
\end{equation}

The simulation result is given in Figure \ref{figure5} (see also Figure \ref{figgg122222}). It is admitted that a force that varies polynomially with time may not be common in practice. Nevertheless, \textit{this example demonstrates the potential that the proposed approach can learn the hidden dynamics of the system together with an independent external force from only one single state trajectory's data.} 
\begin{table}[b!]
	\begin{center}
		\begin{tabular}{|c|c|c|c|c|}
			\hline
			\textbf{\textbf{}}& \textbf{FINS, N=2} & \textbf{FINS\_s, N=2}& \textbf{FINS, N=3}& \textbf{FINS\_s, N=3} \\
			\hline
			Training error& 0.0055 $\pm$ 0.0033  & 0 $\pm$ 0 & 5.4768 $\pm$ 4.6148 & 7.2563 $\pm$ 4.4988\\
			Prediction error& 0.0156 $\pm$ 0.0102  & 0 $\pm$ 0 & 8.9985 $\pm$ 4.6787 & 10.3255 $\pm$ 4.5012\\
			\hline
		\end{tabular}
		\label{polyorder}
  \caption{Effect of the selection of polynomial order on training and prediction error for Example 4}
	\end{center}
\end{table}

To evaluate the effect of polynomial order selection on the learning results, we tested the convergence on training data and evaluated the error on prediction. It is shown in Table 9 that increasing the polynomial order to more than the needed will introduce more modeling error and increase the training error. Since with N = 1, the model is insufficient to converge, we omitted the results in Table 9. In practice, the optimal polynomial order can be determined through an automated search by comparing training errors across candidate polynomial orders. An insufficient polynomial order leads to poor convergence and large training errors, whereas excessively high orders also result in increased training error due to overfitting and numerical instability. In future work, we will develop an automated search algorithm to identify the best-fitting model.



\noindent
\textbf{Example 5}: \textbf{Discovering Lorenz Chaotic System} \\
We select a training time $[0, 10]$ seconds with initial conditions $x(0)=5$, $y(0)=-1$, and $z(0)=2$ to make the dataset. Due to the sensitive dependence on initial
conditions and the complex nonlinear interactions within the chaotic system, we use a high sampling rate. The {\tt ode45} solver gives us a data set $\mathcal{D}_{10}$ having cardinality equal to 13235, so $k(10)=13234$, with \textit{non-uniform} sampling time. We select the highest degree of Hermite polynomials of all variables $x,y,$ and $z$ to be 1, namely $N_{1}=N_{2}=N_{3}=1$.

We select $w(t) \equiv 1, t \in [0,10]$ (see Remark 1) for simulation and apply Algorithm 1, and we set the highest degree of monomials to be 2 for SINDy. The coefficients learned by FINS, FINS\_s, and SINDy are given in Tables 10-12 below. The errors are computed in three intervals: interpolation errors in $[0, 10]$ seconds, extrapolation errors in $[10, 20]$ seconds, and extrapolation errors in $[20, 30]$ seconds, as represented by \textbf{Interpolation, Extrapolation1, Extrapolation2}, respectively, in Table \ref{table54} below.\\

\begin{table}[b!]
	
	\begin{center}
		\begin{tabular}{|c|c|c|c|c|}
			\hline
			\textbf{\textbf{Monomials}}& \textbf{Lorenz system} & \textbf{SINDy}& \textbf{FINS}& \textbf{FINS\_s} \\
			\hline
			1 & 0  & 0 & -0.0005 & 0\\
			x & -10 & -9.9986 & -9.9999 & -9.9999\\
			y & 10 & 9.9991 & 10 & 9.9999\\
			z & 0 & 0 & 0 & 0\\
			xy & 0 & 0 & 0 & 0\\
			xz & 0 & 0 & 0  & 0\\
			yz & 0 & 0 & 0 & 0\\
			\hline
		\end{tabular}
		\label{table51}
  \caption{Coefficients of the monomials of $\dot{x}(t)$ of learned models by SINDy, FINS, and FINS\_s in Example 5 under non-uniform sampling time}
	\end{center}
\end{table}

\begin{table}[t!]
	
	\begin{center}
		\begin{tabular}{|c|c|c|c|c|}
			\hline
			\textbf{\textbf{Monomials}}& \textbf{Lorenz system} & \textbf{SINDy}& \textbf{FINS}& \textbf{FINS\_s} \\
			\hline
			1 & 0  & 0 & -0.0005 & 0\\
			x & 28 & 27.9913 & 28.0002 & 28.0013\\
			y & -1 & -0.9993 & -0.9999  & -1.0003\\
			z & 0 & 0 & 0 & 0 \\
			xy & 0 & 0 & 0 & 0\\
			xz & -1 & -0.9996 & -1 & -1\\
			yz & 0 & 0 & 0 & 0\\
			\hline
		\end{tabular}
		\label{table52}
  \caption{Coefficients of monomials of $\dot{y}(t)$ of learned models by SINDy, FINS, and FINS\_s in Example 5 under non-uniform sampling time}
	\end{center}
\end{table}

\begin{table}[t!]
	
	\begin{center}
		\begin{tabular}{|c|c|c|c|c|}
			\hline
			\textbf{\textbf{Monomials}}& \textbf{Lorenz system} & \textbf{SINDy}& \textbf{FINS}& \textbf{FINS\_s} \\
			\hline
			1 & 0  & 0 & -0.0010 & 0\\
			x & 0 & 0 & 0.0002 & 0\\
			y & 0 & 0 & -0.0001  & 0\\
			z & $-\frac{8}{3}$ & -2.6667 & -2.6666 & -2.6666\\
			xy & 1 & 0.9999 & 1 & 1\\
			xz & 0 & 0 & 0 & 0\\
			yz & 0 & 0 & 0 & 0\\
			\hline
		\end{tabular}
		\label{table53}
  \caption{Coefficients of monomials of $\dot{z}(t)$ of learned models by SINDy, FINS, and FINS\_s in Example 5 under non-unifrom sampling time}
	\end{center}
\end{table}

\begin{table}[t!]
	
	\begin{center}
		\begin{tabular}{|c|c|c|c|}
			\hline
			\textbf{Lorenz System}& \textbf{SINDy} &  \textbf{FINS}& \textbf{FINS\_s}\\
			\hline
			\textbf{Interpolation}  & 0.0101$\pm$0.0054   &  {0.0016$\pm$0.0010} & {\textbf{0.0010$\pm$0.0006}} \\
			\hline
			\textbf{Extrapolation1}  & 0.038 $\pm$  0.016  & {0.008$\pm$ 0.004} & {\textbf{0.003$\pm$ 0.002}} \\
			\hline
			\textbf{Extrapolation2}  & 0.13 $\pm$  0.08 &  {0.04$\pm$ 0.03} & {\textbf{0.02$\pm$ 0.02}} \\
			\hline

		\end{tabular}
  \caption{Comparison of errors of learned models by SINDy, FINS, and FINS\_s in Example 5 under non-uniform sampling time}
		\label{table54}
	\end{center}
\end{table}

\noindent \textbf{Example 6: Discussion on Non-Polynomial Hidden External Input} \\
Consider Example 4 in which the external hidden input is a sinusoidal function of time, i.e., $\delta t^{2}$ is replaced by $0.1 \sin(2t)$. Similar to Subsection 3.2.2, one can see that $f \in L^{2}_{\varphi}(\mathbb{R}^{3}, \mathbb{R}^{2})$. Since the input is \textit{not} polynomial, FINS (or FINS\_s) cannot discover it exactly. We want to see how FINS (or FINS\_s) can handle this case. To do so, we consider a large dataset, i.e., we consider training time $[0,30010]$ seconds to get $k(30010)=11,609,944$ with non-uniform sampling time. We select $N_{1}=N_{2}=N_{3}=3$. Applying Algorithm 1 yields the dynamical system as
\begin{align*}
    \dot{x}&=1.0000y\\
    \dot{y}&=0.0006+0.0186y+0.0076y^{2}+0.0005y^{3}-1.0013x-0.0457xy-0.0001xyt+0.0015xy^{2}\\
    &\quad{}+0.0015xy^{3}-0.0078x^{2}-0.0201x^{2}y-0.0002x^{2}y^{2}-0.0002x^{3}y
\end{align*}
and FINS\_s (where the threshold value of sparsity\footnote{Our simulations for this example (as well as Example 7) show that the dynamics of $\dot{y}$ learned by FINS\_s is highly dependent on the choice of the threshold value, whereas different choices of suitable threshold value of sparsity do not change the learned dynamics of $\dot{x}$.} is chosen as 0.0002) gives
\begin{align*}
			\dot{x}&=1.0000 y  \\
		\dot{y}&=-0.0819+0.0251y-0.0379y^{2}-1.0022x-0.0329xy-0.0030x^{2}-0.0236x^{2}y
\end{align*}
See the following example for further discussion.\\

\begin{table}[t!]
	\begin{center}
		\begin{tabular}{|c|c|c|c|}
			\hline
			\textbf{Lorenz System}& \textbf{SINDy } & \textbf{FINS}& \textbf{FINS\_s}\\
            \hline
			\textbf{\textit{k} = 26k, N = 1} & N/A   & 0.0375$\pm$0.0063  & 0.0403$\pm$0.0103  \\
			\hline
			\textbf{\textit{k} = 260k, N = 1}  & N/A  & 0.3527$\pm$0.0117  & 0.3500$\pm$0.0093   \\
			\hline
			\textbf{\textit{k} = 2.6M, N = 1}  & N/A  & 3.5580$\pm$0.0240  & 3.5850$\pm$0.0157  \\
			\hline
			\textbf{\textit{k} = 26k, N = 2} & 0.0139$\pm$0.0083   & 0.0787$\pm$0.0130  & 0.0805$\pm$0.0113  \\
			\hline
			\textbf{\textit{k} = 260k, N = 2}  & 0.1771$\pm$0.0115  & 0.7625$\pm$0.0108  & 0.7585$\pm$0.0099   \\
			\hline
			\textbf{\textit{k} = 2.6M, N = 2}  & 1.9897$\pm$0.0221  & 7.6285$\pm$0.0414  & 7.6977$\pm$0.0492  \\
			\hline
			\textbf{\textit{k} = 26k, N = 3} &  0.0181$\pm$0.0092  & 0.2366$\pm$0.0138  & 0.2289$\pm$0.0170  \\
			\hline
			\textbf{\textit{k} = 260k, N = 3}  & 0.2367$\pm$0.0173  & 2.2766$\pm$0.0485  & 2.2677$\pm$0.02645 \\
			\hline
			\textbf{\textit{k} = 2.6M, N = 3}  & 2.6706$\pm$0.0255  & 23.3740$\pm$0.1432  & 24.3246$\pm$0.1759  \\
			\hline
			\textbf{\textit{k} = 26k, N = 4} & 0.0251$\pm$0.0058  & 0.7961$\pm$0.0240  &  0.8024$\pm$0.0193 \\
			\hline
			\textbf{\textit{k} = 260k, N = 4}  & 0.2952$\pm$0.0207  & 8.3234$\pm$0.2071   & 8.2459$\pm$0.1931  \\
			\hline
			\textbf{\textit{k} = 2.6M, N = 4}  & 3.3515$\pm$0.0303  & 86.2189$\pm$1.4695  & 87.6410$\pm$1.0147  \\		
			\hline	
		\end{tabular}
  \caption{Comparison of computational time (seconds) (for 25 independent runs) for discovering Lorenz system by SINDy, FINS, and FINS\_s (\textit{if the cardinality of $\mathcal{D}_{t}$ is fixed for the training time}) where \textbf{N} is highest degree of monomials in SINDy and of Hermite polynomials here}
		\label{table91}
	\end{center}
\end{table}

\begin{table}[t!]
	
	\begin{center}
		\begin{tabular}{|c|c|c|}
        \hline
			\textbf{Lorenz System}&  \textbf{FINS}& \textbf{FINS\_s}\\
            \hline
			\textbf{\textit{k} = 26k, N = 1} &  9.1228$\times 10^{-6}$$\pm$3.2566$\times 10^{-5}$   & 5.7904$\times 10^{-5}$$\pm$2.0895$\times 10^{-4}$  \\
			\hline
				\textbf{\textit{k} = 26k, N = 2} &  6.8872$\times 10^{-4}$$\pm$9.7060$\times 10^{-4}$  & 0.0015$\pm$0.0021  \\
			\hline
			\textbf{\textit{k} = 26k, N = 3} &   7.7463$\times 10^{-4}$$\pm$8.1156$\times 10^{-4}$  & 0.0024$\pm$0.0028  \\
				\hline
			\textbf{\textit{k} = 26k, N = 4} &  9.8189$\times 10^{-4}$$\pm$8.3949$\times 10^{-4}$  &  0.0028$\pm$0.0025 \\
					
			\hline
			
					\end{tabular}
  \caption{Comparison of computational time (seconds) for discovering Lorenz system for \textit{incremental learning} (i.e., update time per new datum) of FINS and FINS\_s where \textbf{N} is highest degree of Hermite polynomials}
		\label{table92}
	\end{center}
\end{table}

\noindent \textbf{Example 7: Discussion on Non-Polynomial Vector Field} \\
Consider Mathieu-Hill's equation introduced in Subsection 3.2.2, i.e.,
\begin{equation*}
    \begin{pmatrix}
			\dot{x} \\
			\dot{y} 
		\end{pmatrix}=\begin{pmatrix}
			y \\
			-(\delta+\epsilon  \cos(t))x 
   		\end{pmatrix}
\end{equation*}
where $\delta=1$ and $\epsilon=0.2$. Since the vector field is \textit{not} polynomial, FINS (or FINS\_s) cannot discover it exactly. We want to see how FINS (or FINS\_s) can handle this case. To do so, we consider a large dataset, namely we consider training time $[0,1400]$ seconds with initial conditions $x(0)=20$ and $y(0)=-10$ to get $k(1200)=542,220$ with non-uniform sampling time. We select $N_{1}=N_{2}=N_{3}=3$. Applying Algorithm 1 yields the dynamics as
\begin{align*}
    \dot{x}&=1.0000 y  \\
\dot{y}&=-1.9911+0.0004t+0.0052y-0.0001yt+0.0007y^{2}-0.9844x-0.0001xt-0.0064xy\\
&\quad{}-0.0011x^{2}
\end{align*}
and FINS\_s (where the threshold value of sparsity is chosen to be 0.0001, see also footnote 27) provides
\begin{align*}
			\dot{x}&=1.0000 y  \\
		\dot{y}&=0.0280t+0.0129y+0.0006y^{2}-1.3333x-0.0006x^{2}
\end{align*}
As seen, both FINS and FINS\_s are able to discover $\dot{x}=y$ correctly in Examples 6 and 7. This is because the dynamics of $\dot{x}$ is in the \textit{span}\footnote{The span of finite collection of functions is the set of all linear combinations of them.} of the finite Hermite polynomials considered.
Although FINS and FINS\_s cannot discover a vector field represented by non-polynomials fully, they are able to discover those parts of the vector field which are (is) represented by polynomials if exist(s) and if the highest degrees of Hermite polynomials (here all are 3) are greater than or equal to the corresponding degrees of the polynomial vector field (here for $y$ is 1), see also Example 6.

\begin{table}[t!]
	
	\begin{center}
		\begin{tabular}{|c|c|c|c|c|}
			\hline
			& \textbf{Duffing n=2} & \textbf{Lorenz n=3}& \textbf{Lorenz 96 n=4} & \textbf{Method}\\
			\hline
			\textbf{\textit{k} = 26k, N = 2} & 0.0114$\pm$0.0036  &  0.0139$\pm$0.0083 & 0.0209$\pm$0.0048 & SINDy  \\
			\hline
			\textbf{\textit{k} = 260k, N = 2}  & 0.0896$\pm$0.0042  & 0.1771$\pm$0.0115  & 0.2457$\pm$0.0111 & SINDy \\
			\hline   
			\textbf{\textit{k} = 2.6M, N = 2}  & 0.8026$\pm$0.0058  & 1.9897$\pm$0.0221  & 3.6259$\pm$0.0314 & SINDy \\
			\hline
			\textbf{\textit{k} = 26k, N = 2} & 0.0489$\pm$0.0058  & 0.0787$\pm$0.0130  & 0.3802$\pm$0.0394 & FINS  \\
			\hline
			\textbf{\textit{k} = 260k, N = 2}  & 0.4639$\pm$0.0108  & 0.7625$\pm$0.0108  & 3.8330$\pm$0.0402 & FINS \\
			\hline   
			\textbf{\textit{k} = 2.6M, N = 2}  & 4.7110$\pm$0.0459  & 7.6285$\pm$0.0414  & 37.8838$\pm$0.2629 & FINS \\
			\hline
			\textbf{\textit{k} = 26k, N = 2} & 0.0501$\pm$0.0083  & 0.0805$\pm$0.0113  & 0.3948$\pm$0.0154 & FINS\_s  \\
			\hline
			\textbf{\textit{k} = 260k, N = 2}  & 0.4663$\pm$0.0082  & 0.7585$\pm$0.0099  & 3.9369$\pm$0.0394 & FINS\_s \\
			\hline   
			\textbf{\textit{k} = 2.6M, N = 2}  & 4.7474$\pm$0.0462  & 7.6977$\pm$0.0492  & 36.8969$\pm$0.2515 & FINS\_s \\
			\hline       	
		\end{tabular}
		\label{table10111}
  \caption{Evaluation of computational time (seconds) (\textit{if the cardinality of $\mathcal{D}_{t}$ is fixed for the training time})  w.r.t. the dimension of the states, i.e., \textbf{n}, (for 25 independent runs) where \textbf{N} is highest degrees of monomials in SINDy and of Hermite polynomials}
	\end{center}
\end{table}

\begin{table}[t!]
	\scriptsize
	\begin{center}
		\begin{tabular}{|c|c|c|c|}
			\hline
			\textbf{}& \textbf{Duffing n=2} & \textbf{Lorenz n=3}& \textbf{Lorenz 96 n=4}\\
			\hline
			\textbf{FINS} & 5.8375$\times 10^{-4}$$\pm$7.8622$\times 10^{-4}$  & 6.8872$\times 10^{-4}$$\pm$9.7060$\times 10^{-4}$  & 5.1372$\times 10^{-4}$$\pm$4.4352$\times 10^{-4}$  \\
					\hline 
                    \textbf{FINS\_s} & 0.0014$\pm$0.0020  & 0.0015$\pm$0.0021  & 0.0081$\pm$0.0053  \\
					\hline
		\end{tabular}
		\label{table102}
  \caption{Evaluation of computational time (seconds) for \textit{incremental learning} (i.e., update time per new datum)  w.r.t. the dimension of the states, i.e., \textbf{n}, where $k$ = 26k and \textbf{N=2}}
	\end{center}
\end{table}

\begin{figure}[t!]
	\begin{center}		\centerline{\includegraphics[scale = 0.7]{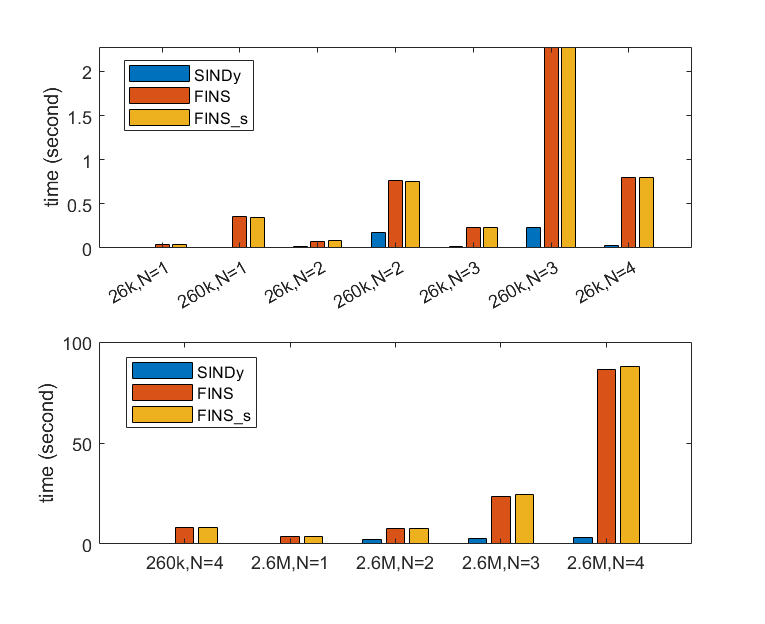}}
		\caption{Bar plot of Table 14 }
		\label{figure3january11}
	\end{center}
\end{figure}

\subsection{Evaluation of Computational Time}

\begin{figure}[t!]
	\begin{center}		\centerline{\includegraphics[scale = 0.5]{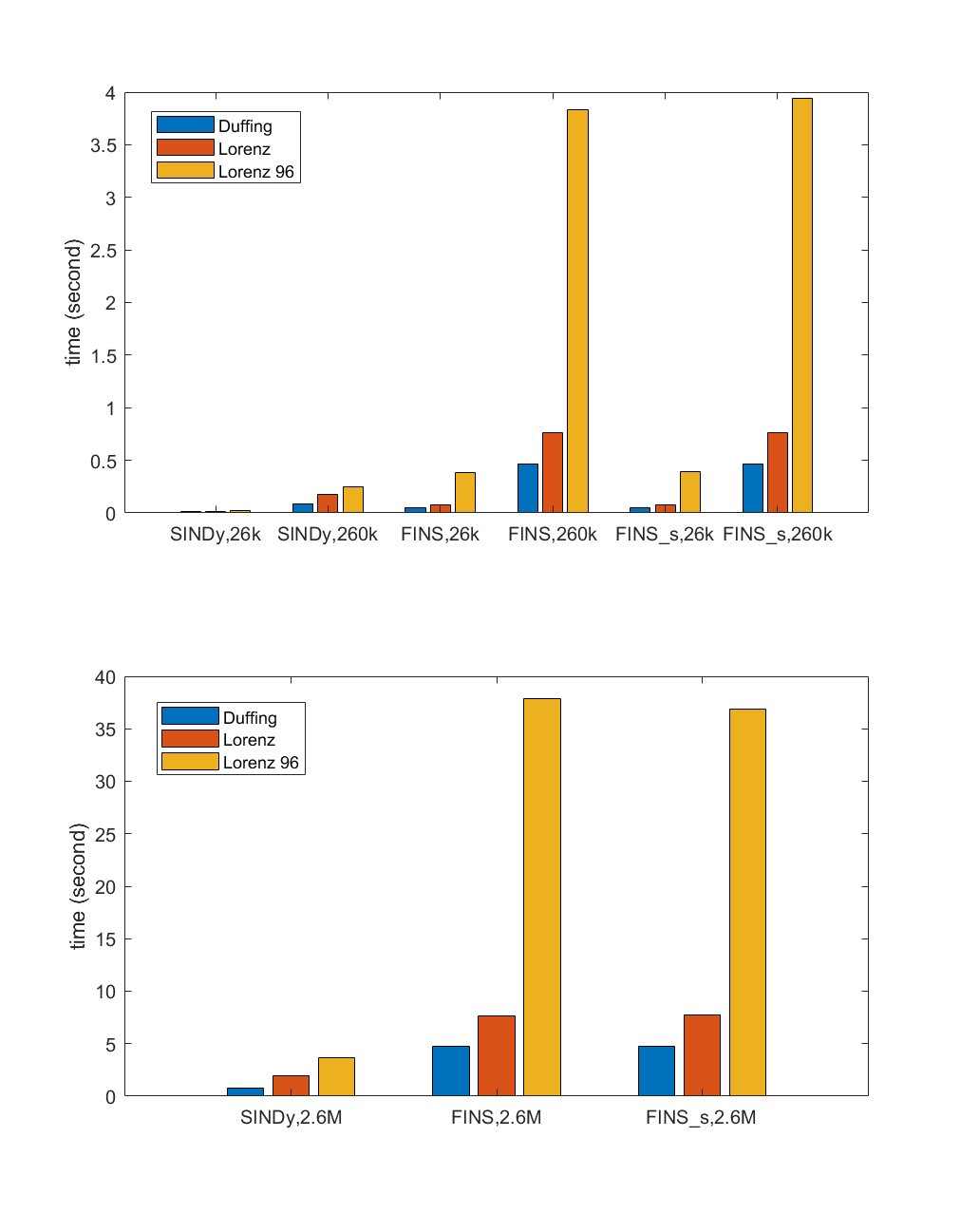}}
		\caption{Bar plot of Table 16 }
		\label{figure3january22}
	\end{center}
\end{figure}

With the Duffing ($n=2$), Lorenz ($n=3$), and Lorenz 96 ($n=4$) systems as examples, the upper bound of computational time\footnote{We ran the simulation with x64-based
processor Intel(R) Core(TM) i7-1165G7 CPU 2.80~GHz and 15.7~GB
RAM usable. \textit{To improve computational time, we used calling a function to compute $\mathcal{H}(\underline{\textbf{x}})$ in our MATLAB codes}. Note that since $n$ equations in (\ref{corollarylinearalgebraicequation}) are calculated in \textit{parallel}, their computational time is equal to that of one of them (similarly for calculating computational time of  $P_{k}^{l}, V_{k}^{l}, \hdots, l=1,\hdots,n$).} are shown in Tables 14 and 16 with respect to scalability. Regarding Table 14, the numbers of parameters used in SINDy for $N=2,3,4,$ (where $N$ is the higehst degree of monomials) are 10, 20, and 35, respectively, whereas the numbers of parameters used in FINS and FINS\_s for $N=1,2,3,4,$ (where $N$ is the highest degree of Hermite polynomials) are 8, 27, 64, and 125, respectively. In Table 16, the numbers of parameters used in SINDy for Duffing, Lorenz, and Lorenz 96 systems are 6, 10, and 15, respectively, while the number of parameters used in FINS (or FINS\_s) for Duffing, Lorenz, and Lorenz 96 systems are 9, 27, and 81, respectively. The bar plots of Tables 14 and 16 are provided in Figures \ref{figure3january11} and \ref{figure3january22}, respectively. 

In the case where the cardinality of $\mathcal{D}_{t}$ is fixed for the training time (see Algorithm 1), as seen from Tables 14 and 16, the computational time of SINDy is faster than that of the proposed method here (considering the fact that the number of parameters used in SINDy are lower). However, due to the incremental learning capability of Algorithm 1, the update time per new data is \textit{significantly low} (see Tables 15 and 17). Since incremental learning capabilities of existing methods for ODE discovery have not been shown in the literature, the results in Tables 15 and 17 are not comparable with existing results.


\subsection{Evaluation and Comparison with the State-of-the-Art Network-Based Algorithms}
To evaluate the status-quo performance of data-driven discovery of differential equations with recent neural-network and other network-based methods rather than algorithms compared above, two state-of-the-art algorithms were evaluated: Symbolic Regression via Neural Networks (SymNN) (\cite{symbolicRgeressionviaNN2023}) and ODENet (\cite{ODENet}).

\vspace{1mm}
\noindent \textbf{SymNN}: We use the original author-created codes for SymNN to evaluate and compare SymNN's performance with an example adopted from \cite{symbolicRgeressionviaNN2023}. As shown in \cite{symbolicRgeressionviaNN2023}, multiple random trajectories are used for training in most examples evaluated except low-order systems like the Lorenz system. We evaluated SymNN using a single trajectory data without noise. We adopt Takens–Bogdanov normal form equation\footnote{We select Takens–Bogdanov normal form equation because its vector field is in polynomial form and the result in \cite{symbolicRgeressionviaNN2023} provides minimum approximate relative-RMSE (see Table III in \cite{symbolicRgeressionviaNN2023}).} for which SymNN does not converge to the underlying equation if a single trajectory of 1 second is used, whereas the proposed method converges. If 4~seconds of data is used, SymNN can converge. A summary of the comparison is provided below. The simualted Takens–Bogdanov equation is 
\begin{align*}
    \dot{x}&=y \\
    \dot{y}&= x^2 + xy +1.5y -4.41
\end{align*}

With one trajectory of data sampled from Takens–Bogdanov equation for $t \in [0,4]$ seconds including 1002 uniformly-sampled data\footnote{The authors in \cite{symbolicRgeressionviaNN2023} used 1002 points in their code.} without noise, the best equation SymNN identifies in 12,800 epochs with 3-decimal accuracy is: 
\begin{align*}
    \dot{x}&=y \\
    \dot{y}&= x^2 + xy + 1.5y -4.333 
\end{align*}

It is worth mentioning that the above results are consistent with those reported in the SymNN paper and remain the same under different random initial conditions. The observed non-convergence with shorter trajectories and the error in the learned equation by SymNN are likely due to its use of fractional representations with small integers for estimating coefficients. When the true underlying coefficients cannot be accurately represented in this form, the algorithm tends to either fail to converge or produce biased results.  With the same dataset, for $N_{1}=N_{2}=2$ and with 0.001 accuracy, FINS gives the dynamics as
\begin{align*}
    \dot{x}&=y \\    \dot{y}&=-4.409+1.499y+0.999xy+x^{2}
\end{align*}
while FINS\_s yields
\begin{align*}
    \dot{x}&=y \\    
    \dot{y}&=-4.410+1.500y+xy+x^{2}
\end{align*}
It can be seen that FINS\_s obtained the exact solutions for the same example.

\vspace{1mm}

\noindent \textbf{ODENet}: We use the original code associated with \cite{ODENet}\footnote{The code is available at https://github.com/cam1681/ODENet.}. We tested the algorithm on Duffing equation in Example 1 under non-uniform sampling time data without noise. With 10,000 epochs, ODENet provides the following dynamics: 
\begin{align*}
    \dot{x}&=-2.5577x + 1.1427y \\
    \dot{y}&= -59.98x -4.1964y -1.2147 x^3
\end{align*}
Fine tuning of method parameters was attempted but unsuccessful to achieve convergent results. 
Admittedly, with proper fine tuning, ODENet may converge to the actual solution. However, this indicates that gradient-based methods suffer from a robustness issue with the step size and other hyperparameters and need a case-by-case fine tuning for each equation and each dataset with different length and sampling rate.

\textbf{Computational time}: We also measured the training time for SymNN and ODENet. For SymNN to learn Takens–Bogdanov equation, with 12,800 training epochs, the total training time is 7,749.08 seconds, while for ODENet, with 10,000 epochs for training on Duffing equation, the training time is 1,002.30 seconds. These results indicate that network-based methods using gradient-descent-based optimization algorithms are at least $10^5$ times slower than the proposed approach (see also Tables 14 and 16). This significant difference in computational efficiency highlights the potential of the proposed method for \textit{real-time} applications, whereas the computational burden of the compared methods renders them impractical in such scenarios.

\section{Conclusions and Discussion}	
In this paper, data-driven learning of nonlinear dynamical systems represented by ODEs is reformulated, and a new interpretable formulation based on Functional Analysis and Operator Theory is presented. The training cost function in this method is a continuous sum, in contrast to the discrete sum in the existing machine learning methods, which allows developing incremental learning algorithms. The proposed method has several capabilities and advantages: it can discover the exact underlying physics for systems represented by finite degree polynomials only from one single state trajectory's data; it can decouple and learn both the unknown dynamics and unknown external input simultaneously; and the results are valid for time-varying dynamical systems under forced and/or unforced systems. Finally, numerical examples have been provided to show the effectiveness of the learning capability of the proposed method, even with noisy data. 

Despite advantages of the proposed method shown, it has limitations, like any other methods, that need to be improved. In this regard, some future research subjects are provided: 1) while we show via examples that it is possible to increase the accuracy by increasing the number of data, the mathematical proof of influence over finite time requires further research, 2) we have shown via a numerical example that the performance of FINS seems to be robust to noise. Nevertheless, providing mathematical proof and comparing with existing methods needs further investigation, 3) addressing the curse of dimensionality of the proposed method is another subject for future research, 4) theoretical studies on the convergence properties
of Algorithm 1 under infinite data needs to be further investigated, and 5) while we use Hermite polynomials in the results, utilizing other orthogonal functions and comparing with the results here opens a subject for future work.




\acks{


\textbf{Disclosure of Funding:}
Yongzhi Qu acknowledges the partial support by the Department of Commerce of the United States under contract No. 70NANB20H175 and 70NANB22H109, and Seyyed Shaho Alaviani acknowledges the McKnight postdoctoral fellow support from August 2022 to July 2023 from the University of Minnesota.

	\textbf{Authors Contributions:}	``S. Sh. Alaviani `contributed' Conceptualization, Methodology, Formal Analysis, Investigation, Software, Validation, Writing; Y. Qu `contributed' Conceptualization, Software, Validation, Formal Analysis, Investigation, Writing, and Supervision; G. W. Vogl `contributed' Software, Validation, Formal Analysis, Investigation, and Writing.''
	
	\textbf{Competing Interests:} The authors declare no competing interests.}
	
	\textbf{Disclaimer:} Certain commercial equipment, instruments, or materials are identified in this paper in order to specify the experimental procedure adequately. Such identification is not intended to imply recommendation or endorsement by the National Institute of Standards and Technology, nor is it intended to imply that the materials or equipment identified are necessarily the best available for the purpose. This material is declared a work of the U.S. Government and is not subject to copyright protection in the United States. Approved for public release; distribution is unlimited.



\appendix
\section{}

\noindent
\textbf{Proof of Lemma 1:}	We require to show that $$T(\alpha f_{1}+\beta f_{2})(t)=\alpha T(f_{1})(t)+\beta T(f_{2})(t)$$ for all $\alpha \in \mathbb{R}$ and $\beta \in \mathbb{R}$. Therefore, we obtain
\begin{align*}
	T(\alpha f_{1}+\beta f_{2})(t) &=\int_{t_{0}}^{t} [\alpha f_{1}(s,x(s))+\beta f_{2}(s,x(s))]ds \\
	&=\int_{t_{0}}^{t} \alpha f_{1}(s,x(s)) ds+\int_{t_{0}}^{t} \beta f_{2}(s,x(s)) ds \\
	&=\alpha \int_{t_{0}}^{t}  f_{1}(s,x(s)) ds+\beta \int_{t_{0}}^{t}  f_{2}(s,x(s)) ds\\
	&=\alpha T(f_{1})(t)+\beta T(f_{2})(t)
\end{align*}
which implies that the operator $T(f)$ defined in (\ref{operatorT}) is linear. Thus the proof of Lemma 1 is complete. 

\section{}

\noindent
\textbf{Proof of Preposition 1:}  Hermite polynomials (see Definition 2 for details) form an orthogonal basis in $L^{2}_{\psi}(\mathbb{R},\mathbb{R})$ (with inner product defined in (\ref{weightinhermitedefinition})). Hence, \textbf{any} function $\bar{f} \in L^{2}_{\psi}(\mathbb{R},\mathbb{R})$ can be written as an infinite sum of the form $$\bar{f}(y)=\sum_{j=0}^{\infty} \bar{c}_{j} H_{j}(y)$$ where $\bar{c}_{j} \in \mathbb{R}$ are coefficients, and $H_{j}(y)$ are defined in (\ref{hermitepolyderivation}) for $j=0,1,\hdots$.

Similarly, \textit{any} function $f \in L^{2}_{\varphi}(\mathbb{R} \times \mathbb{R}^{n},\mathbb{R}^{n})$ (with inner product defined in (\ref{normf})) can be written as 
\begin{equation}\label{4444444444}
	f(\underline{\textbf{x}})=col\{f^{1}(\underline{\textbf{x}}),\hdots,f^{n}(\underline{\textbf{x}})\} 
\end{equation}
where 
\begin{align*}
	f^{l}(\underline{\textbf{x}}) &=\sum_{i_{1}=0}^{\infty} \sum_{i_{2}=0}^{\infty} \hdots \sum_{i_{n+1}=0}^{\infty} (c^{l}_{i_{1} i_{2} \hdots i_{n+1}}  H_{i_{1}}(x_{1}) H_{i_{2}}(x_{2})  \hdots H_{i_{n}}(x_{n}) H_{i_{n+1}}(t) ) .
\end{align*}
Now we approximate a function $f$ in (\ref{4444444444}) by Hermite polynomials of \textit{finite degree}, i.e.,
\begin{equation}\label{rrrrrrrr}
	\hat{f}(\underline{\textbf{x}})=col\{\hat{f}^{1}(\underline{\textbf{x}}),\hdots,\hat{f}^{n}(\underline{\textbf{x}})\}
\end{equation}
where
\begin{align*}
	\hat{f}^{l}(\underline{\textbf{x}}) &=\sum_{i_{1}=0}^{N_{1}} \sum_{i_{2}=0}^{N_{2}} \hdots \sum_{i_{n+1}=0}^{N_{n+1}} (c^{l}_{i_{1} i_{2} \hdots i_{n+1}}  H_{i_{1}}(x_{1}) H_{i_{2}}(x_{2})  \hdots H_{i_{n}}(x_{n}) H_{i_{n+1}}(t) ) 
\end{align*}
in which $N_{j} \in \mathbb{N},j=1,\hdots,n+1.$ (\ref{rrrrrrrr}) can be rearranged in a more compact form as
\begin{equation}\label{5555555555555}
	\hat{f}(\underline{\textbf{x}})=C \mathcal{H}(\underline{\textbf{x}})
\end{equation}
where $C \in \mathbb{R}^{n \times (N_{1}+1)(N_{2}+1) \hdots (N_{n+1}+1)}$, $\mathcal{H}(\underline{\textbf{x}}) \in \mathbb{R}^{(N_{1}+1)(N_{2}+1)\hdots (N_{n+1}+1)}$, and $C$ and $\mathcal{H}$ are defined in (\ref{matrixC}) and (\ref{matrixH}), respectively. Substitution of (\ref{5555555555555}) into (\ref{minimizationformulation}) yields
\begin{equation}\label{6666666666666}
	\begin{aligned}
		& \underset{C_{k} }{\text{min}}
		& & \Vert \int_{t_{0}}^{\bar{t}} C_{k} \mathcal{H}(x(s),s) ds -x(\bar{t})+x(t_{0})+\int_{t_{0}}^{\bar{t}} \mathcal{F}(s)ds \Vert_{L^{2}_{w}} \\
		& \text{s.t.}
		& & \{(t_{j},x(t_{j}),\mathcal{F}(t_{j}))\}_{j=0}^{k} \subseteq \mathcal{D}_{t}, 
	\end{aligned}
\end{equation} 	
where $C_{k} \in \mathbb{R}^{n \times (N_{1}+1)(N_{2}+1) \hdots (N_{n+1}+1)}$ and $k \in \mathbb{N} \cup \{0\}$. Since $L^{2}_{w}([t_{0},t_{k}],\mathbb{R}^{n})$ (and in general $L^{2}_{w}(\mathbb{R},\mathbb{R}^{n})$) is a Hilbert space, it is a strictly convex space \cite[Ch. 2]{stricthilbert}. Therefore, (\ref{6666666666666}) and (\ref{77777777777}) are equivalent\footnote{By \textit{equivalent} we mean that the optimal solutions of both optimization problems are equal.}. This completes the proof of Preposition 1.

\section{}

\noindent
\textbf{Proof of Theorem 1:} We have from (\ref{normxtruncated}) and (\ref{77777777777}) that
\begin{align}
	&\Vert \int_{t_{0}}^{\bar{t}} C_{k} \mathcal{H}(x(s),s) ds -x(\bar{t})+x(t_{0}) +\int_{t_{0}}^{\bar{t}} \mathcal{F}(s)ds \Vert_{L^{2}_{w}}^{2}  
	 \nonumber \\
	&=\int_{t_{0}}^{t_{k}} w(z) \Vert \int_{t_{0}}^{z} C_{k} \mathcal{H}(x(s),s) ds- x(z)+x(t_{0}) +\int_{t_{0}}^{z} \mathcal{F}(s)ds \Vert_{2}^{2} dz \nonumber \\
	&=\int_{t_{0}}^{t_{k}} w(z)[\int_{t_{0}}^{z} C_{k} \mathcal{H}(x(s),s) ds- x(z)+x(t_{0})+ \int_{t_{0}}^{z} \mathcal{F}(s)ds]^{T}  [\int_{t_{0}}^{z} C_{k} \mathcal{H}(x(s),s) ds- x(z) \nonumber \\
	&\quad{}+x(t_{0})+\int_{t_{0}}^{z} \mathcal{F}(s)ds] dz \nonumber \\
	&=\int_{t_{0}}^{t_{k}} w(z) [\int_{t_{0}}^{z} C_{k} \mathcal{H}(x(s),s) ds]^{T}  [\int_{t_{0}}^{z} C_{k} \mathcal{H}(x(s),s) ds] dz \label{part111}\\
	&\quad{}  +\int_{t_{0}}^{t_{k}} w(z) [\int_{t_{0}}^{z} C_{k} \mathcal{H}(x(s),s) ds]^{T}[x(t_{0})-x(z)+ \int_{t_{0}}^{z} \mathcal{F}(s)ds] dz \label{part222}\\
	&\quad{} +\int_{t_{0}}^{t_{k}} w(z) [x(t_{0})-x(z)+\int_{t_{0}}^{z} \mathcal{F}(s)ds]^{T}  [\int_{t_{0}}^{z} C_{k} \mathcal{H}(x(s),s) ds] dz \label{part333}\\
	& \quad{}+\int_{t_{0}}^{t_{k}} w(z) [x(t_{0})-x(z)+\int_{t_{0}}^{z} \mathcal{F}(s)ds]^{T}  [x(t_{0})-x(z)+\int_{t_{0}}^{z} \mathcal{F}(s)ds] dz. \label{part444}
\end{align}
Now we obtain from (\ref{part111}) and (\ref{mathcalKdefinition}) that
\begin{align}
	\int_{t_{0}}^{t_{k}} w(z) [\int_{t_{0}}^{z} C_{k} \mathcal{H}(x(s),s) ds]^{T}  [\int_{t_{0}}^{z} C_{k} \mathcal{H}(x(s),s) ds] dz  
	&=\int_{t_{0}}^{t_{k}} w(z) \Vert C_{k} \mathcal{K}(z) \Vert_{2}^{2} dz \nonumber \\
	&=\int_{t_{0}}^{t_{k}} (w(z) \sum_{l=1}^{n} \Vert C_{k}^{l} \mathcal{K}(z) \Vert_{2}^{2}) dz \nonumber \\
	&=\int_{t_{0}}^{t_{k}} (w(z) \sum_{l=1}^{n} C_{k}^{l} \mathcal{K}(z)\mathcal{K}^{T}(z) C_{k}^{l^{T}}) dz \nonumber \\
	&= \sum_{l=1}^{n} C_{k}^{l} (\int_{t_{0}}^{t_{k}} w(z) \mathcal{K}(z) \mathcal{K}^{T}(z) dz) C_{k}^{l^{T}} \nonumber \\
	&=\mathcal{C}^{T}_{k} Q_{k} \mathcal{C}_{k}. \label{derivationforQ}
\end{align}
where $Q_{k}$ is defined in (\ref{Qkdefinition}). We obtain from (\ref{part222}), (\ref{part333}), and (\ref{mathcalKdefinition}) that
\begin{align}
	&\int_{t_{0}}^{t_{k}} w(z) [\int_{t_{0}}^{z} C_{k} \mathcal{H}(x(s),s) ds]^{T}[x(t_{0})-x(z)  +\int_{t_{0}}^{z} \mathcal{F}(s)ds] dz \nonumber\\
	&\quad{} +\int_{t_{0}}^{t_{k}} w(z) [x(t_{0})-x(z)+\int_{t_{0}}^{z} \mathcal{F}(s)ds]^{T} [\int_{t_{0}}^{z} C_{k} \mathcal{H}(x(s),s) ds] dz \nonumber \\
	&=2 \int_{t_{0}}^{t_{k}} w(z)[\int_{t_{0}}^{z} C_{k} \mathcal{H}(x(s),s) ds]^{T}[x(t_{0})-x(z)  +\int_{t_{0}}^{z} \mathcal{F}(s)ds] dz \nonumber \\
	&=2 \int_{t_{0}}^{t_{k}} w(z) [C_{k} \mathcal{K}(z)]^{T} [x(t_{0})-x(z)+\int_{t_{0}}^{z} \mathcal{F}(s)ds] dz \nonumber \\
	&=2 \int_{t_{0}}^{t_{k}} (w(z) \sum_{l=1}^{n} C_{k}^{l} \mathcal{K}(z) [x_{l}(t_{0})-x_{l}(z)+\int_{t_{0}}^{z} \mathcal{F}_{l}(s)ds]) dz \nonumber \\
	&= \sum_{l=1}^{n} C_{k}^{l} [2 \int_{t_{0}}^{t_{k}} w(z) \mathcal{K}(z)(x_{l}(t_{0})-x_{l}(z)+\int_{t_{0}}^{z} \mathcal{F}_{l}(s)ds)dz] \nonumber \\
	&=P^{T}_{k} \mathcal{C}_{k} \label{derivationforP}
\end{align}
where $P_{k}$ is defined in (\ref{Pkdefinition}). We also have from (\ref{part444}) that 
\begin{align}
	&\int_{t_{0}}^{t_{k}} w(z) [x(t_{0})-x(z)+\int_{t_{0}}^{z} \mathcal{F}(s)ds]^{T}  [x(t_{0})-x(z)+\int_{t_{0}}^{z} \mathcal{F}(s)ds]dz \nonumber\\
	&=\int_{t_{0}}^{t_{k}} w(z) \Vert x(t_{0})-x(z)+\int_{t_{0}}^{z} \mathcal{F}(s)ds \Vert_{2}^{2} dz=U_{k} \label{derivationforU}		
\end{align}
where $U_{k}$ is defined in (\ref{Ukdefinition}). Therefore, we have from (\ref{part111})-(\ref{derivationforU}) that solving (\ref{77777777777}) reduces to (\ref{888888888}). Hence, the proof of Theorem 1 is complete.

\section{}

\noindent
\textbf{Proof of Corollary 1:} Since $Q_{k}$, for some $k$, is a positive semi-definite matrix, the cost function in (\ref{888888888}) is a convex function. Since (\ref{888888888}) is an \textit{unconstrained} minimization, we obtain through the \textit{first-order optimality condition} that an optimal solution $\mathcal{C}_{k}^{*}$ satisfies
$$\nabla (\mathcal{C}_{k}^{*^{T}} Q_{k} \mathcal{C}_{k}^{*} + P^{T}_{k} \mathcal{C}_{k}^{*}+U_{k})=\textbf{0}_{n(N_{1}+1) \hdots (N_{n+1}+1)}$$
which implies (\ref{corollarylinearalgebraicequation}) since $Q_{k}$ is both block-diagonal and symmetric. Therefore, the proof of Corollary 1 is complete.

\section{}

\noindent
\textbf{Proof of Corollary 2:} We obtain from (\ref{Qktildedefinition}) that
\begin{align}
	\tilde{Q}_{k+1} &=\int_{t_{0}}^{t_{k+1}} w(z) \mathcal{K}(z) \mathcal{K}^{T}(z) dz =\int_{t_{0}}^{t_{k}} w(z) \mathcal{K}(z) \mathcal{K}^{T}(z) dz +\int_{t_{k}}^{t_{k+1}} w(z) \mathcal{K}(z) \mathcal{K}^{T}(z) dz \nonumber \\
	&=\tilde{Q}_{k}+\int_{t_{k}}^{t_{k+1}} w(z) \mathcal{K}(z) \mathcal{K}^{T}(z) dz \label{qqqq1}
\end{align}
where $\mathcal{K}(z)$ is defined in (\ref{mathcalKdefinition}). We have that 
\begin{align}
	\mathcal{K}(z)&=\int_{t_{0}}^{z} \mathcal{H}(x(s),s) ds =\int_{t_{0}}^{t_{k}} \mathcal{H}(x(s),s) ds+\int_{t_{k}}^{z} \mathcal{H}(x(s),s) ds =\mathcal{K}(t_{k})+\int_{t_{k}}^{z} \mathcal{H}(x(s),s) ds. \label{qqqq2}
\end{align}
Substituting (\ref{qqqq2}) for (\ref{qqqq1}) yields
\begin{align}
	\tilde{Q}_{k+1} &=\tilde{Q}_{k}+\int_{t_{k}}^{t_{k+1}} w(z) [\mathcal{K}(t_{k})+\int_{t_{k}}^{z} \mathcal{H}(x(s),s) ds]  [\mathcal{K}(t_{k}) +\int_{t_{k}}^{z} \mathcal{H}(x(s),s) ds]^{T} dz \label{QQQQ} 
\end{align}
where $\tilde{Q}_{0}=\textbf{0}$ in which $\textbf{0}$ is the zero matrix with appropriate dimension. From the data set $\mathcal{D}_{t}$ defined in (\ref{dataset}) and (\ref{integralapproximation}), we can approximate $\mathcal{K}(t_{k})$ as
\begin{align*}
	\mathcal{K}(t_{k}) &\approx \sum_{j=1}^{k} (t_{j}-t_{j-1})(\frac{\mathcal{H}(x(t_{j}),t_{j})+\mathcal{H}(x(t_{j-1}),t_{j-1})}{2})=\mathcal{K}(t_{k-1})+(t_{k}-t_{k-1}) M_{k}
\end{align*}
where 
\begin{equation}
	M_{k}:=\frac{\mathcal{H}(x(t_{k}),t_{k})+\mathcal{H}(x(t_{k-1}),t_{k-1})}{2},
\end{equation} 
and $\mathcal{K}(t_{0})=\textbf{0}$ (in which $\textbf{0}$ is the zero vector with appropriate dimension). Similarly, we can approximate the following integration
\begin{equation}\label{Q111}
	\int_{t_{k}}^{z} \mathcal{H}(x(s),s) ds \approx M_{k+1}(z-t_{k}), \quad{} t_{k} \leq z \leq t_{k+1}.
\end{equation}
Substituting (\ref{Q111}) for (\ref{QQQQ}) yields
\begin{align}
	\tilde{Q}_{k+1} &=\tilde{Q}_{k}+\int_{t_{k}}^{t_{k+1}} w(z) [\mathcal{K}(t_{k})+\int_{t_{k}}^{z} \mathcal{H}(x(s),s) ds]  [\mathcal{K}(t_{k}) +\int_{t_{k}}^{z} \mathcal{H}(x(s),s) ds]^{T} dz \nonumber \\
	&=\tilde{Q}_{k}+\int_{t_{k}}^{t_{k+1}} w(z) [\mathcal{K}(t_{k})+M_{k+1}(z-t_{k})]  [\mathcal{K}(t_{k}) +M_{k+1}(z-t_{k})]^{T} dz \nonumber \\
	&=\tilde{Q}_{k}+\mathcal{K}(t_{k})\mathcal{K}^{T}(t_{k}) \int_{t_{k}}^{t_{k+1}} w(z) dz +\mathcal{K}(t_{k}) M^{T}_{k+1}  [\int_{t_{k}}^{t_{k+1}} w(z) (z-t_{k}) dz] \nonumber \\
	&\quad{}+[\int_{t_{k}}^{t_{k+1}} w(z) (z-t_{k}) dz] M_{k+1} \mathcal{K}^{T}(t_{k})  + [\int_{t_{k}}^{t_{k+1}} w(z) (z-t_{k})^{2} dz] M_{k+1}M^{T}_{k+1}. \label{wwwwwwwww}
\end{align}
It should be noted that since $w(t)$ is known, the integrals $\int_{t_{k}}^{t_{k+1}} w(z) dz$, $\int_{t_{k}}^{t_{k+1}} w(z) (z-t_{k}) dz$, and $\int_{t_{k}}^{t_{k+1}} w(z) (z-t_{k})^{2} dz$ in (\ref{wwwwwwwww}) can be calculated either analytically or by the integral approximation (\ref{integralapproximation}).

Now we obtain
\begin{align}
	P^{l}_{k+1}&=2 \int_{t_{0}}^{t_{k+1}} w(z)  \mathcal{K}(z) [x_{l}(t_{0})-x_{l}(z)+\int_{t_{0}}^{z} \mathcal{F}_{l}(s)ds] dz \nonumber \\
	&=2 \int_{t_{0}}^{t_{k}} w(z) \mathcal{K}(z) [x_{l}(t_{0})-x_{l}(z)+\int_{t_{0}}^{z} \mathcal{F}_{l}(s)ds] dz \nonumber \\
	&\quad{} + 2\int_{t_{k}}^{t_{k+1}} w(z) \mathcal{K}(z) [x_{l}(t_{0})-x_{l}(z)+\int_{t_{0}}^{z} \mathcal{F}_{l}(s)ds] dz \nonumber \\
	&=P^{l}_{k}+2\int_{t_{k}}^{t_{k+1}} w(z) \mathcal{K}(z) [x_{l}(t_{0})-x_{l}(z) +\int_{t_{0}}^{z} \mathcal{F}_{l}(s)ds] dz. \label{uuuu}
\end{align} 

Substituting (\ref{qqqq2}) for (\ref{uuuu}) yields
\begin{align}
	P^{l}_{k+1}&=P^{l}_{k}+2 \int_{t_{k}}^{t_{k+1}} w(z) [\mathcal{K}(t_{k}) +\int_{t_{k}}^{z} \mathcal{H}(x(s),s) ds] [x_{l}(t_{0})-x_{l}(z)+\int_{t_{0}}^{z} \mathcal{F}_{l}(s)ds] dz \nonumber \\
	&=P^{l}_{k}+2 \mathcal{K}(t_{k}) \int_{t_{k}}^{t_{k+1}} w(z) [x_{l}(t_{0})-x_{l}(z) +\int_{t_{0}}^{z} \mathcal{F}_{l}(s)ds] dz \nonumber \\
	&\quad{} + 2 \int_{t_{k}}^{t_{k+1}} w(z) [\int_{t_{k}}^{z} \mathcal{H}(x(s),s) ds]  [x_{l}(t_{0})-x_{l}(z) +\int_{t_{0}}^{z} \mathcal{F}_{l}(s)ds] dz. \label{ffgghh}
\end{align}
Substituting (\ref{Q111}) for (\ref{ffgghh}) implies
\begin{align}
	P^{l}_{k+1}&=P^{l}_{k}+2 \mathcal{K}(t_{k}) \int_{t_{k}}^{t_{k+1}} w(z) [x_{l}(t_{0})-x_{l}(z) +\int_{t_{0}}^{z} \mathcal{F}_{l}(s)ds] dz   \nonumber \\
	 &\quad{}+2 M_{k+1} \int_{t_{k}}^{t_{k+1}} w(z) [z-t_{k}][x_{l}(t_{0}) -x_{l}(z)+\int_{t_{0}}^{z} \mathcal{F}_{l}(s)ds] dz \nonumber \\
	&=P^{l}_{k}+2 \mathcal{K}(t_{k}) \int_{t_{k}}^{t_{k+1}} w(z) [x_{l}(t_{0})-x_{l}(z)]dz \nonumber \\
	&\quad{} + 	2 \mathcal{K}(t_{k}) \int_{t_{k}}^{t_{k+1}} w(z)[\int_{t_{0}}^{t_{k}} \mathcal{F}_{l}(s)ds+\int_{t_{k}}^{z}  \mathcal{F}_{l}(s)ds]dz \nonumber \\
	&\quad{}+ 2 M_{k+1} \int_{t_{k}}^{t_{k+1}} w(z) [z-t_{k}][x_{l}(t_{0})-x_{l}(z)]dz \nonumber \\
	&\quad{}+ 2 M_{k+1} \int_{t_{k}}^{t_{k+1}} w(z)[z-t_{k}] [\int_{t_{0}}^{t_{k}} \mathcal{F}_{l}(s)ds +\int_{t_{k}}^{z}  \mathcal{F}_{l}(s)ds]dz. \label{ppllkk} 
\end{align}	

Now we have that
\begin{align}\label{thetallllll}
	\theta^{l}(t_{k}) :=\int_{t_{0}}^{t_{k}} \mathcal{F}_{l}(s)ds &\approx \sum_{j=1}^{k} (t_{j}-t_{j-1}) \frac{\mathcal{F}_{l}(t_{j})+\mathcal{F}_{l}(t_{j-1})}{2} \\
	&= \theta^{l}(t_{k-1})+(t_{k}-t_{k-1}) \psi^{l}(t_{k})
\end{align}	
where
\begin{equation}\label{saiiiiiiiikkkk}
	\psi^{l}(t_{k}):=\frac{\mathcal{F}_{l}(t_{k})+\mathcal{F}_{l}(t_{k-1})}{2}.
\end{equation}	

From (\ref{integralapproximation}), (\ref{thetallllll}), and (\ref{ppllkk}), we obtain	
\begin{align}
	P^{l}_{k+1} &=P^{l}_{k}+2 \mathcal{K}(t_{k}) \int_{t_{k}}^{t_{k+1}} w(z) [x_{l}(t_{0})-x_{l}(z)]dz \nonumber \\
	&\quad{} + 	2 \mathcal{K}(t_{k}) \int_{t_{k}}^{t_{k+1}} w(z)[\theta^{l}(t_{k})+\psi^{l}(t_{k+1})(z-t_{k})]dz \nonumber \\
	&\quad{}+ 2 M_{k+1} \int_{t_{k}}^{t_{k+1}} w(z) [z-t_{k}][x_{l}(t_{0})-x_{l}(z)]dz \nonumber \\
	&\quad{}+ 2 M_{k+1} \int_{t_{k}}^{t_{k+1}} w(z) [z-t_{k}] [\theta^{l}(t_{k}) +\psi^{l}(t_{k+1})(z-t_{k})]dz. \label{ppllkk111111111111} 
\end{align}	

From  (\ref{integralapproximation}) and (\ref{ppllkk111111111111}), we obtain
\begin{align*}
	P^{l}_{k+1}&=P^{l}_{k}+2(t_{k+1}-t_{k}) \mathcal{K}(t_{k}) V^{l}_{k+1} +2 \mathcal{K}(t_{k}) \theta^{l}(t_{k}) \int_{t_{k}}^{t_{k+1}} w(z)dz \\
	&\quad{}+ 2 \mathcal{K}(t_{k})\psi^{l}(t_{k+1}) \int_{t_{k}}^{t_{k+1}} w(z)(z-t_{k})dz	+2(t_{k+1}-t_{k}) M_{k+1} G^{l}_{k+1}\\
	&\quad{}+ 2M_{k+1} \theta^{l}(t_{k}) \int_{t_{k}}^{t_{k+1}} w(z)(z-t_{k})dz  + 2M_{k+1}\psi^{l}(t_{k+1}) \int_{t_{k}}^{t_{k+1}} w(z)(z-t_{k})^{2}dz
\end{align*}
in which
\begin{align*}
	V^{l}_{k+1}&:=\frac{w(t_{k+1})[x_{l}(t_{0})-x_{l}(t_{k+1})]}{2}+\frac{w(t_{k})[x_{l}(t_{0})-x_{l}(t_{k})]}{2}, \\
	G^{l}_{k+1}&:=\frac{w(t_{k+1})(t_{k+1}-t_{k})[x_{l}(t_{0})-x_{l}(t_{k+1})]}{2}.
\end{align*}	
Similar to the above, since $w(t)$ is known, the integrals $\int_{t_{k}}^{t_{k+1}} w(z) dz$, $\int_{t_{k}}^{t_{k+1}} w(z) (z-t_{k}) dz$, and $\int_{t_{k}}^{t_{k+1}} w(z) (z-t_{k})^{2} dz$ in (\ref{wwwwwwwww}) can be calculated either analytically or by the integral approximation (\ref{integralapproximation}). Thus the proof of Corollary 2 is complete.

\section{}

\textbf{Updates of $\tilde{Q}_{k}$ and $P_{k}^{l}$ in Algorithm 1:}	\\	 

$M_{k} \gets \frac{\mathcal{H}(x(t_{k}),t_{k})+\mathcal{H}(x(t_{k-1}),t_{k-1})}{2}$

$V^{l}_{k} \gets \frac{w(t_{k})(x_{l}(t_{0})-x_{l}(t_{k}))+w(t_{k-1})(x_{l}(t_{0})-x_{l}(t_{k-1}))}{2}$

$G^{l}_{k} \gets \frac{w(t_{k})(t_{k}-t_{k-1})(x_{l}(t_{0})-x_{l}(t_{k}))}{2}$

$\mathcal{K}_{k} \gets \mathcal{K}_{k-1}+(t_{k}-t_{k-1})M_{k}$

$\psi_{k}^{l} \gets \frac{\mathcal{F}_{l}(t_{k})+\mathcal{F}_{l}(t_{k-1})}{2}$

$\theta^{l}_{k} \gets \theta^{l}_{k-1}+(t_{k}-t_{k-1})\psi_{k}^{l}$
\begin{align*}
	\tilde{Q}_{k} \gets \tilde{Q}_{k-1} &+ \mathcal{K}_{k-1} \mathcal{K}^{T}_{k-1} \int_{t_{k-1}}^{t_{k}} w(z) dz+\mathcal{K}_{k-1}M_{k}^{T} \int_{t_{k-1}}^{t_{k}} w(z)(z-t_{k-1}) dz\\
	&+[\int_{t_{k-1}}^{t_{k}} w(z)(z-t_{k-1}) dz] M_{k}\mathcal{K}_{k-1}^{T}+[\int_{t_{k-1}}^{t_{k}} w(z)(z-t_{k-1})^{2} dz] M_{k} M_{k}^{T}\\			
	P_{k}^{l} \gets P_{k-1}^{l} &+2(t_{k}-t_{k-1}) \mathcal{K}_{k-1} V_{k}^{l} +2(t_{k}-t_{k-1}) M_{k}G_{k}^{l} +2 \mathcal{K}_{k-1}\theta^{l}_{k-1} \int_{t_{k-1}}^{t_{k}} w(z)dz \\
	&  +2\mathcal{K}_{k-1}\psi^{l}_{k} \int_{t_{k-1}}^{t_{k}} w(z)(z-t_{k-1})dz +2M_{k}\theta^{l}_{k-1} \int_{t_{k-1}}^{t_{k}} w(z)(z-t_{k-1})dz \\
	& +2M_{k} \psi^{l}_{k} \int_{t_{k-1}}^{t_{k}} w(z)(z-t_{k-1})^{2} dz 
\end{align*}

\vskip 0.2in
\bibliography{sample11}

\end{document}